\documentclass[sigconf]{acmart}

\copyrightyear{2026}
\acmYear{2026}
\setcopyright{cc}
\setcctype{by}
\acmConference[WWW '26]{Proceedings of the ACM Web Conference 2026}{April 13--17, 2026}{Dubai, United Arab Emirates}
\acmBooktitle{Proceedings of the ACM Web Conference 2026 (WWW '26), April 13--17, 2026, Dubai, United Arab Emirates}
\acmPrice{}
    \acmDOI{10.1145/3774904.3792637}
\acmISBN{979-8-4007-2307-0/2026/04}

\AtBeginDocument{%
  }

\usepackage{amsmath,amssymb,amsfonts}
\usepackage{graphicx}
\usepackage{textcomp}
\usepackage{xcolor}
\usepackage{multicol}
\usepackage{amsthm}
\usepackage[linesnumbered,ruled]{algorithm2e}
\usepackage{caption}
\usepackage{subfigure}
\usepackage{url}
\usepackage{multirow}
\usepackage{float}
\usepackage{enumitem}
\usepackage{booktabs}

\usepackage{xcolor}
\usepackage{caption}

\begin{document}

\title{Amortized Predictability-aware Training Framework for Time Series Forecasting and Classification}

\settopmatter{authorsperrow=4}
\author{Xu Zhang}
\orcid{0009-0006-5317-2422}
% \authornote{This work was conducted when Xu Zhang was interning at Ant Group. This work was supported by Ant Group Research Intern Program.}

\affiliation{%
  \institution{Shanghai Key Laboratory of Data Science, College of Computer Science and Artificial Intelligence\\ Fudan University}
  \city{Shanghai}
  \country{China}
}
\email{xuzhang22@m.fudan.edu.cn}

\author{Peng Wang}
\authornote{Peng Wang is the corresponding author.}
\affiliation{%
  \institution{Shanghai Key Laboratory of Data Science, College of Computer Science and Artificial Intelligence\\ Fudan University}
  \city{Shanghai}
  \country{China}
}
\email{pengwang5@fudan.edu.cn}

\author{Yichen Li}
\affiliation{%
  \institution{Department of Electrical and Computer Engineering \\ University of British Columbia (UBC)}
  \city{Vancouver}
  \country{Canada}
}
\email{yli262@student.ubc.ca}

\author{Wei Wang}
\affiliation{%
  \institution{Shanghai Key Laboratory of Data Science, College of Computer Science and Artificial Intelligence\\Fudan University}
  \city{Shanghai}
  \country{China}
  % \streetaddress{P.O. Box 1212}
  % \country{USA}
}
\email{weiwang1@fudan.edu.cn}

% \author{Wei Wang}
% % \authornote{Both authors contributed equally to this research.}
% % \orcid{1234-5678-9012}
% % \author{G.K.M. Tobin}
% % \email{webmaster@marysville-ohio.com}
% \affiliation{%
%   \institution{Shanghai Key Laboratory of Data Science, School of Computer Science\\ Fudan University}
%   \city{Shanghai}
%   \country{China}
%   % \streetaddress{P.O. Box 1212}
%   % \city{Dublin}
%   % \state{Ohio}
%   % \country{USA}
%   % \postcode{43017-6221}
% }
% \email{weiwang1@fudan.edu.cn}

% \renewcommand{\shortauthors}{Xu Zhang et al.}
\renewcommand{\shortauthors}{Xu Zhang, PengWang, Yichen Li, and Wei Wang}

\begin{abstract}

Time series data are prone to noise in various domains, and training samples may contain low-predictability patterns that deviate from the normal data distribution, leading to training instability or convergence to poor local minima. Therefore, mitigating the adverse effects of low-predictability samples is crucial for time series analysis tasks such as time series forecasting (TSF) and time series classification (TSC).
While many deep learning models have achieved promising performance, few consider how to identify and penalize low-predictability samples to improve model performance from the training perspective. 
To fill this gap, we propose a general Amortized Predictability-aware Training Framework (APTF) for both TSF and TSC. APTF introduces two key designs that enable the model to focus on high-predictability samples while still learning appropriately from low-predictability ones: (i) a Hierarchical Predictability-aware Loss (HPL) that dynamically identifies low-predictability samples and progressively expands their loss penalty as training evolves, and (ii) an amortization model that mitigates predictability estimation errors caused by model bias, further enhancing HPL's effectiveness.
The code is available at \url{https://github.com/Meteor-Stars/APTF}.
\end{abstract}




\maketitle
% \newcommand\webconfavailabilityurl{https://doi.org/xxxx}
% \ifdefempty{\webconfavailabilityurl}{}{
% \begingroup\small\noindent\raggedright\textbf{Resource Availability:}\\
% % please change the following context to include multiple artifacts if necessary, including data, models, code, etc.
% The source code of this paper has been made publicly available at \url{\webconfavailabilityurl}.
% \endgroup
% }

\section{Introduction}

Time series (TS) data is ubiquitous in web applications, such as ECG signals in EHRs~\cite{sarkar2020self}, fund sales on Alipay~\cite{zhang2024self}, and IoT sensor data~\cite{zhang2025global}.
By analyzing temporal patterns, time series forecasting (TSF) and time series classification (TSC) can help make decisions, e.g., predict future trends for inventory planning~\cite{zhang2025multi} or identify events occurring within the current temporal patterns for healthcare monitoring~\cite{forestier2018surgical}. 
As a result, these techniques have become increasingly important in modern web technologies~\cite{huang2022semi,zhang2025global}.

TS data are prone to noise. For instance, industrial sensors may suffer recording errors due to wear and environmental factors~\cite{nunes2023challenges}; ECG signals can be corrupted by electromyographic interference in medical applications~\cite{pandit2017noise}; and stock sequences in the financial sector often exhibit noise caused by market fluctuations~\cite{diebold2013correlation}. 
As a result, TS samples frequently contain low-predictability patterns~\cite{cho2022wavebound}, which deviate from the normal data distribution and exhibit higher noise levels. 
Moreover, deep neural networks have been demonstrated to have a high capacity to fit noisy data during training, which will further result in poor performance and generalizationon the test data~\cite{zhang2021understanding}. 
Importantly, this phenomenon does not change with the choice of training optimizations or network architectures~\cite{jiang2018mentornet,zhang2021understanding}.
\textbf{The high loss values of low-predictability samples can dominate gradient updates, causing training oscillations and trapping the model in poor local minima~\cite{cho2022wavebound}.}
Therefore, mitigating the adverse impact of low-predictability samples is essential for improving model convergence.

Although deep-learning-based methods perform well in TSF~\cite{kim2024self,wangtimexer} and TSC~\cite{cheng2023formertime,tang2020omni}, they typically treat all training samples equally and ignore how to identify and handle low-predictability samples to achieve better convergence. 
Ideally, such samples should incur smaller losses, enabling the model to learn from them without overemphasis while paying more attention to high-predictability samples. 
In the computer vision domain, Co-teaching~\cite{han2018co} removes noisy samples based on loss values and shows strong robustness to label noise. 
This leads to the question: \textit{Can Co-teaching be adapted to time series tasks, and what modifications are required?}

We show that adopting Co-teaching for time series is non-trivial, as directly applying the vanilla Co-teaching framework may not improve TS models and can even degrade their performance.
Co-teaching is designed for settings with extremely noisy labels, where such samples are explicitly harmful and can be safely discarded. 
In contrast, our setting does not involve extreme noise. 
\textbf{Low predictability samples may still provide useful information, and discarding them can waste knowledge and harm performance}. 
The key challenge is to learn from these samples while reducing the risk of overfitting their noise. 
To this end, we propose an Amortized Predictability-aware Training Framework (APTF) for TSF and TSC tasks.

Noisy and low-predictability samples both deviate from the data distribution. 
Given Co-teaching's great success in reducing noise by discarding samples with high-loss values, we similarly use high-loss values to identify low-predictability samples.  
However, deep models have been proven to first memorize clean (high-predictability) samples, and gradually adapt to noisy (low-predictability) samples as training proceeds~\cite{arpit2017closer}. 
This suggests that loss values can reliably reflect sample predictability in early training, but become unreliable as training progresses due to overfitting, which is a phenomenon we term the \textit{predictability evolution} issue.

This motivates us to propose a Predictability-aware Training Framework (PTF) that dynamically identifies low-predictability samples as training evolves. 
The core of PTF is the Hierarchical Predictability-aware Loss (HPL). It groups samples into multiple buckets based on their predictability from high to low  (determined by loss magnitude), 
and penalizes the gradients of low-predictability samples by assigning progressively smaller loss weights.
Then, the hierarchical bucketing strategy across training stages further addresses the \textit{predictability evolution} issue. As a result, HPL emphasizes highly predictable samples while down-weighting low-predictability ones, reducing the risk of overfitting to noise while still learning appropriately from low-predictability ones.

In PTF, the process of finding low-predictability samples is dynamically evolving by a single model itself.

However, the model's decisions can be overconfident and biased~\cite{freund1999short,balcan2006agnostic}, leading to predictability estimation errors where highly predictable samples are misestimated as low-predictability ones, vice versa.
When using only a single model, accumulated estimation errors may cause over-penalization of highly predictable samples and degrade performance. 
To alleviate this, we introduce an amortization model to reduce predictability estimation error and further improve PTF, resulting in the Amortized Predictability-aware Training Framework (APTF). 
Our contributions are as follows:

\begin{itemize}
\item We propose a general Amortized Predictability-aware Training Framework (APTF) for TSF and TSC tasks. 
To our knowledge, this is the first framework that improves training for both tasks by dynamically identifying and penalizing low-predictability samples throughout training.

\item In APTF, we propose a Hierarchical Predictability-aware Loss (HPL) to address the \textit{predictability evolution} issue. 
HPL enables the model to focus on high-predictability samples while still learning appropriately from low-predictability ones, with a reduced risk of overfitting to their noise.
An amortization model is further designed to alleviate predictability estimation errors, improving the effectiveness of HPL.

\item We conduct extensive experiments on 11 TSF datasets and 128 TSC datasets from the UCR archive. 
Results show that APTF improves convergence accuracy and generalization across eleven TSF and five TSC models, reducing prediction errors by an average of 2\%--9.79\% on long-term and 3\%--15.36\% on short-term TSF tasks.

\end{itemize}

\section{Related work}
\label{sec:related_work}
\textbf{Time series forecasting (TSF).}
Initially, Recurrent Neural Networks (RNNs)~\cite{jordan1997serial,cho2014learning} were popular for TSF, but their recurrent structure limits scalability. Temporal Convolutional Networks (TCNs)~\cite{sen2019think}, based on causal and dilated convolutions, improve parallelism and are widely used in TSF~\cite{sen2019think}. However, both RNNs and CNNs struggle to capture long-term dependencies due to structural or receptive-field limitations. To address this, many methods adapt Transformers~\cite{vaswani2017attention} to model both short- and long-term dependencies, such as Autoformer~\cite{wu2021autoformer}, Scaleformer~\cite{shabani2022scaleformer}, and MLF~\cite{zhang2025multi}. Recently, MLP-based linear models have also shown strong performance~\cite{zhou2022film,chen2023tsmixer,DBLP:journals/tmlr/DasKLMSY23}, including NLinear~\cite{zeng2023transformers_linear}, N-HiTS~\cite{challu2022n_Nhits}, and LSINet~\cite{zhang2025lightweight}. In addition, CNN-based models~\cite{luo2024moderntcn} and cross-attention methods~\cite{kim2024self} achieve a favorable balance between efficiency and accuracy.

\textbf{Time series classification (TSC).}
Recently, deep learning has gained popularity in TSC due to its strong feature extraction ability, high accuracy, and scalability~\cite{Ismail_Muller_2019}. One-dimensional CNNs (1D-CNNs) are widely used for TSC because of their effectiveness in capturing local temporal features~\cite{tang2020rethinking}, leading to methods such as InceptionTime~\cite{ismail2020inceptiontime}, OS-CNN~\cite{tang2020omni}, FCN, and ResNet~\cite{wang2017time}. More recently, FormerTime~\cite{cheng2023formertime} integrates 1D-CNNs with Transformers to jointly model local and global time-series features for TSC.

\textbf{Methods for predictability-aware learning.}
Some studies mitigate the impact of low-predictability samples by modifying loss computation. For example, Co-teaching~\cite{han2018co} discards high-loss samples to handle extremely noisy labels, thereby enhancing the model's robustness.
Nevertheless, such samples can still be meaningful in our setting, motivating the proposed Hierarchical Predictability-aware Loss (HPL). Self-paced learning~\cite{kumar2010self} instead learns from low-loss samples first and gradually includes harder ones, while adaptive robust loss~\cite{barron2019general} suppresses outliers via learnable loss parameters. In TSF, WaveBound~\cite{cho2022wavebound} avoids overfitting unpredictable steps by adjusting gradients when prediction errors are lower than estimated bounds.

%\vspace{-0.1cm}
\section{Amortized predictability-aware training framework (APTF)}

\begin{figure*}[bt]
% %\vspace{-0.3cm}
% \setlength{\abovecaptionskip}{0.1cm}
\centerline{\includegraphics[width=0.8\linewidth]{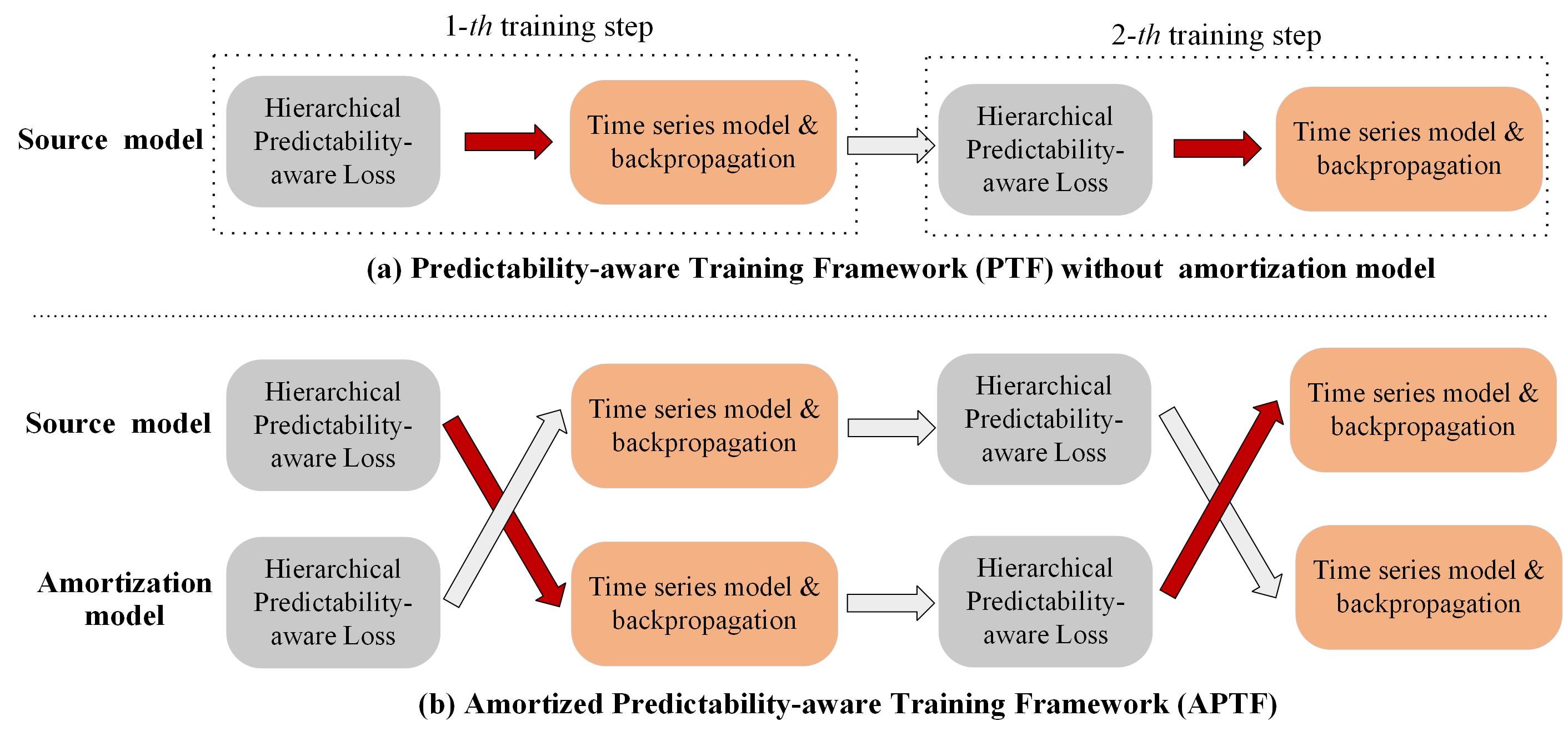} }
\vspace{-0.2cm}
\caption{The proposed general training framework APTF for time series forecasting and classification. }
\label{fig:model}
\vspace{-0.1cm}
\label{fig:frame_work}
\end{figure*}

APTF is illustrated in Figure~\ref{fig:frame_work}(b). Its core component is the Hierarchical Predictability-aware Loss (HPL), detailed in Algorithm~\ref{alg:algorithm2} and Figure~\ref{fig:evolution_aware}. HPL adopts a hierarchical bucketing strategy to penalize the loss of low-predictability samples as training evolves, allowing the model to learn from them while reducing the risk of overfitting their noise. By introducing an amortization model, predictability estimation bias is first absorbed and then passed to the source model with a one-step delay (red arrows in Figure~\ref{fig:frame_work}(b)), mitigating the misestimation of sample predictability caused by overconfidence. Notably, HPL alone already yields substantial performance gains, while the amortization model further enhances its effectiveness.

\subsection{Hierarchical predictability-aware loss (HPL)}
\subsubsection{\textbf{Low-predictability to high-predictability buckets} }
We define the training set as $\mathcal{D}=\{(X_1,Y_1),...,(X_N,Y_N)\}$, in which $\mathcal{X}_N = [x_1, x_2,
\ldots, x_t] \in \mathbb{R}^{t \times v}$ with length $t$ for all $v$ time variables. 
$Y_N$ varies depending on the downstream task. For example, in the in the forecasting task, $Y_N=[x_{t+1},x_{t+2},...,x_{t+m}]$ are the consecutive time values of $X_i$, while in the classification task, $Y_N \in {1,...,C}$ are $C$ class labels. 
During training, we optimize the TS model $f(X,\Theta)$ through the basic function:
 \begin{equation}
 % %\vspace{-0.1cm}
\label{equ:loss_func}
\arg\min_{\Theta} \sum_{(X_i,Y_i) \in \mathcal{D}}  \mathcal{L}\left( f(X_i,\Theta),Y_i\right) 
% %\vspace{-0.05cm}
\end{equation}
where $\mathcal{L}$ denotes Mean Squared Error (MSE) Loss for forecasting or cross-entropy loss for classification. 
For clarity, we illustrate the following method using the forecasting task as an example.

Low-predictability samples typically incur larger losses, which can cause training oscillations and convergence to poor local minima. Thus, it is necessary to identify these samples and regularize them to prevent the model from overfocusing on them and avoid harming the model's learning capacity from high-predictability samples.
Since loss values are effective indicators of noisy samples~\cite{han2018co}, we use high-loss values to identify low-predictability samples in our setting.

We first uniformly divide samples into $K$ buckets ${B_1,\dots,B_K}$ according to ascending loss values.
\textbf{Buckets with larger indices contain samples with higher loss and thus lower predictability}, e.g., $B_K$ contains the least predictable samples.
Each bucket has $N_k^B=\lfloor \frac{N}{K} \rfloor$ samples, where $N$ is the batch size and $\lfloor\cdot\rfloor$ denotes the floor operation. After filling the first $K-1$ buckets evenly, all remaining samples are assigned to the last bucket.
As shown in Figure~\ref{fig:evolution_aware}(a), when the samples can be evenly divided, $Len(B_1)=Len(B_2)=\cdots=Len(B_K)$, where $Len(\cdot)$ denotes the number of samples in a bucket.
The example is shown in Figure~\ref{fig:evolution_aware}(a), when $N$ is divisible by $K$, $Len(B_1)=\cdots=Len(B_K)$, where $Len(\cdot)$ denotes the number of samples in a bucket.

\subsubsection{\textbf{Loss weights for paying more attention to high predictability samples} }

The basic buckets $\{B_1,...,B_K\}$ group samples by predictability.
To reduce the adverse influence of low-predictability samples, we assign smaller loss weights to higher-index buckets and larger weights to lower ones, i.e., $\{W_1>W_2>...>W_K\}$.
This amplifies the gradients of high-predictability samples, guiding the model to focus more on them.
We empirically observe that uniformly decreasing weights yield good performance in practice.
The resulting objective is termed the basic predictability-aware loss, as shown in Algorithm~\ref{alg:algorithm1} and Figure~\ref{fig:evolution_aware}(a).

\begin{algorithm}[bt]
\footnotesize 
\caption{Basic \textit{Predictability\_Aware\_Loss}.}
	\label{alg:algorithm1}
	
	\KwIn{
 \begin{enumerate}
 \item $K$ buckets $\{B_1, B_2, ..., B_K\}$, each bucket containing the loss value of samples with different predictability levels. 
 \item $K$ weights $\{W_1, W_2. ..., W_K\}$ for penalizing the loss values in different buckets. 
 \item Initialized final loss value $\mathcal{L}=0$.
 \end{enumerate}}
        \KwOut{The loss value $\mathcal{L}$.}
	\BlankLine

		\For{$j=1$ to $K$}{
                $\mathcal{L}_j$=Average($B_j \cdot W_j$)\\
                $\mathcal{L}=\mathcal{L}+\mathcal{L}_j$\\}
            \Return $\mathcal{L}$
            % \Return {$\frac{\mathcal{L}}{K}$}
\end{algorithm}

\begin{algorithm}[bt]
\caption{Amortized hierarchical predictability-aware loss at the $S$-th training stage}
\label{alg:algorithm2}
\footnotesize
\KwIn{
\begin{enumerate}
    \item Hierarchical bucket groups $\{\mathcal{B}_1, \ldots, \mathcal{B}_{S}\}$ and weight groups $\{\mathcal{W}_1, \ldots, \mathcal{W}_S\}$ from \textbf{source model} up to the $S$-th training stage.
    \item Bucket groups $\{\widetilde{\mathcal{B}}_1, \ldots, \widetilde{\mathcal{B}}_S\}$ and weight groups $\{\widetilde{\mathcal{W}}_1, \ldots, \widetilde{\mathcal{W}}_S\}$ from \textbf{amortization model} up to the $S$-th training stage.
    \item Initialized loss values $\mathcal{L}=0$ and $\widetilde{\mathcal{L}}=0$ for source and amortization models.
    \item Basic function \textit{Predictability\_Aware\_Loss($\cdot$)} in Algorithm~\ref{alg:algorithm1}.
    \item Whether to use amortization model: \textit{amortization=True}.
    \item Total number of training stages $G$.
\end{enumerate}
}
\KwOut{Loss $\mathcal{L}$ and $\widetilde{\mathcal{L}}$ for optimizing source and amortization models.}
\BlankLine
\For{$i=1$ to $S$}{
    \eIf{\textit{amortization}=True}{
        $\mathcal{L} = \mathcal{L} + \textit{Predictability\_Aware\_Loss}(\widetilde{\mathcal{B}}_i, \widetilde{\mathcal{W}}_i)$ \\
        $\widetilde{\mathcal{L}} = \widetilde{\mathcal{L}} + \textit{Predictability\_Aware\_Loss}(\mathcal{B}_i, \mathcal{W}_i)$
    }{
        $\mathcal{L} = \mathcal{L} + \textit{Predictability\_Aware\_Loss}(\mathcal{B}_i, \mathcal{W}_i)$ \\
        $\widetilde{\mathcal{L}} = 0$
    }
}
\Return $\mathcal{L}/G$, \ $\widetilde{\mathcal{L}}/G$
\end{algorithm}

\subsubsection{\textbf{Predictability-aware loss with predictability evolution}}
The basic predictability-aware loss in Algorithm~\ref{alg:algorithm1} adopts a fixed number of buckets throughout training, resulting in a fixed gradient penalty range for low-predictability samples.
It fails to account for the adverse effects of overfitting during training, which may \textbf{lead to the low-predictability sample not being penalized because it was assigned to a lower-index bucket due to its lower loss value resulting from overfitting}.
This phenomenon, referred to as the \textit{predictability evolution} issue, indicates that loss values effectively estimate sample predictability in early training stages but become less accurate as training progresses.

To address this, we first uniformly divide the training into $S$ stages at the training epoch interval $\varepsilon$ to dynamically reduce the number of buckets as the stage increases. 
The next training stage will be activated when the indicator $tag=1$:
% %%\vspace{-0.1cm}
\begin{equation}
% %%\vspace{-0.2cm}
\label{equ:cal_global_randomness}
    tag=
    \begin{cases}
    1 &  \text{if} \ \ \text{current epoch} \ \% \ \varepsilon=0 \\
    0 &  \text{otherwise} \\
    \end{cases}
%%\vspace{-0.1cm}
\end{equation}

The decreased number of buckets at different training stages is illustrated in Figure~\ref{fig:evolution_aware}(b), which shows an example of three training stages. 
As training progresses, samples originally in high-weight buckets are gradually reassigned to lower-weight ones by reducing the total number of buckets and removing the highest loss weights.
This progressively enlarges the range of penalized low-predictability samples.
The updated buckets and weights are then fed into Algorithm~\ref{alg:algorithm1} to compute the training loss for optimization.

\subsubsection{\textbf{Hierarchical Predictability-aware Loss with predictability evolution}}
\begin{figure*}[h]
% %\vspace{-0.3cm}
% \setlength{\abovecaptionskip}{0.1cm}
\centerline{\includegraphics[width=1\linewidth]{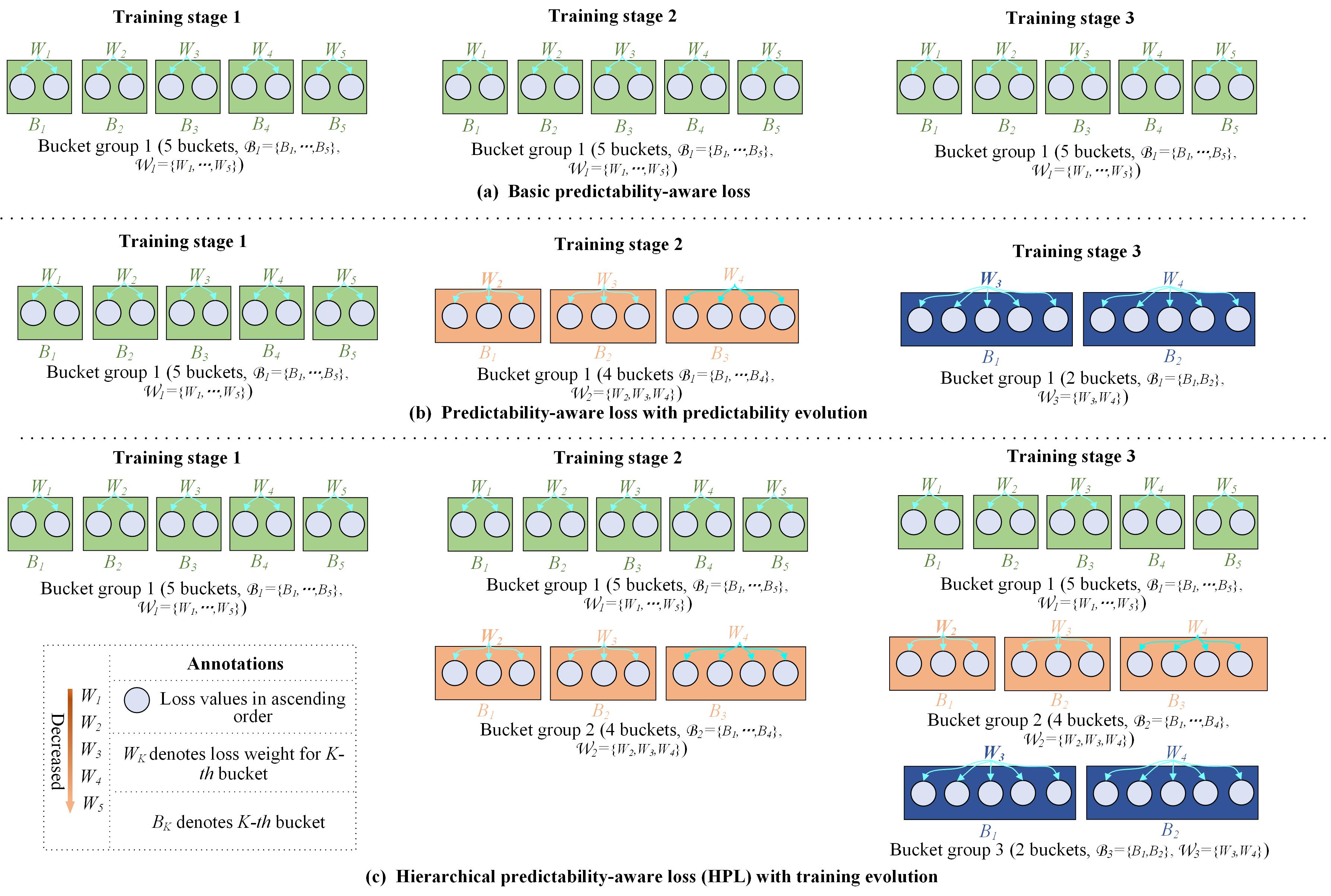} }
\vspace{-0.2cm}
\caption{Illustration of different bucketing strategies for computing the predictability-aware loss.
\textbf{(a)} A fixed number of buckets is used throughout training, ignoring the \textit{predictability evolution} issue.
\textbf{(b)} The number of buckets decreases as training progresses, addressing \textit{predictability evolution} but discarding previous bucket partitions.
\textbf{(c)} Hierarchical buckets consider both \textit{predictability evolution} and previous bucket partitioning strategies.}
\label{fig:evolution_aware}
\vspace{-0.1cm}
\end{figure*}

Although the design in Figure~\ref{fig:evolution_aware}(b) mitigates the \textit{predictability evolution} issue, it still has some limitations. 
\textbf{First}, the loss weights used to penalize samples vary across different training stages, which can lead to significant changes in the model's gradient direction, potentially causing unstable training or falling into poor local optima. 
\textbf{Second}, \textbf{as the training stage evolves, more samples will be penalized with lower weights and the gradient strength becomes weaker, which can prevent some highly predictable samples from obtaining sufficient gradients to learn due to their low loss values}, thereby harming model performance.

To address these limitations, we further propose the \textbf{Hierarchical Predictability-aware Loss (HPL)}, illustrated in Figure~\ref{fig:evolution_aware}(c).
We define a \textit{bucket group} as ${\mathcal{B}_1,\ldots,\mathcal{B}_G}$, where each group contains several buckets, e.g., $\mathcal{B}_1=\{B_1,...,B_K\}$, with corresponding weight groups ${\mathcal{W}_1,\ldots,\mathcal{W}_G}$.
As training progresses, the number of bucket groups gradually increases, while newly added groups contain fewer buckets, leading to more samples per bucket.
Meanwhile, lower weights are assigned to progressively expand the penalty on low-predictability samples, following the same principle as before.
The ``hierarchy'' indicates that at the $S$-th training stage ($S>1$), the number of bucket groups is always greater than one.
As shown in Figure~\ref{fig:evolution_aware}(c), \textbf{stage 1} contains a single bucket group with 5 buckets, whereas \textbf{stage 2} contains two bucket groups with 5 and 3 buckets, respectively.
In contrast, without HPL, each training stage has only one bucket group (Figure~\ref{fig:evolution_aware}(b)).

As training progresses, we incorporate previous bucket partition strategies into the loss computation and average the losses across different bucket groups.
\textbf{On one hand}, this avoids abrupt gradient changes that could cause training oscillations or poor local optima.
\textbf{On the other hand}, it compensates the loss of highly predictable samples, thereby amplifying their gradients to mitigate the risks of harming the model's learning ability due to over-penalization. 
This achieves a balance where high-predictability samples are sufficiently fitted while low-predictability ones are appropriately penalized.
For $S$ training stages, the Hierarchical Predictability-aware Loss is computed as shown in Algorithm~\ref{alg:algorithm2} (with \textit{amortization} set to False).

\subsection{Amortized hierarchical predictability-aware loss}
In HPL, low-predictability samples are dynamically identified by a single model, which may introduce bias due to overconfidence and lead to misidentification of sample predictability~\cite{freund1999short,balcan2006agnostic}.
Inspired by co-training~\cite{blum1998combining}, we introduce an amortization model to reduce predictability estimation error caused by model bias.
By mutually estimating sample predictability, the bias of the source model is first amortized (absorbed) by the amortization model and then propagated back after a one-step delay, as shown by the red arrows in Figure~\ref{fig:frame_work}(b).
This reduces the risk of over-penalizing highly predictable samples due to misidentification and further improves HPL performance.
The pseudocode of the amortized hierarchical predictability-aware loss is provided in Algorithm~\ref{alg:algorithm2}.

\section{Experiments and results}
\subsection{Experimental settings}
% \textbf{\textbf{Datasets}
\textbf{Datasets for short-term TSF.} We introduce fund sales datasets from~\cite{zhang2025multi} to validate the effectiveness of our method. 
\textbf{The reason is that fund sales exhibit significant volatility under the influence of complex factors such as policy, public opinion, and market trends, making them more prone to low-predictability patterns.} 
They are divided into three groups based on the holding period for comprehensive experiment evaluation, called Fund1 (containing 66 fund sales datasets), Fund2 (107 fund sales datasets), and Fund3 (133 fund sales datasets). Details can be found in Appendix Section~\ref{sec:fund_discuss} and Figure~\ref{fig:series_comp_appen}.

\textbf{Datasets for long-term TSF.} We evaluate the performance of our proposed framework on eight popular datasets, including Exchange, Weather, Electricity, Traffic, and four ETT datasets (ETTh1, ETTh2, ETTm1, ETTm2). 
These datasets involve applications in industrial machines,
energy, and weather domains. 
% The statistics of them are shown in Table~\ref{tab:dataset_stat}, covering a range of time steps and variables. 
They have been widely employed in the literature for multivariate forecasting tasks~\cite{nie2022time_patchformer,wu2021autoformer,wangtimemixer}. 
Following previous works, each dataset is split into training, validation, and testing sets in a 7:1:2 ratio. 
In short-term forecasting, we merged the training, validation, and test sets across all fund datasets for model training and evaluation because they share similar TS patterns.

\textbf{Datasets for TSC.} We validate the effectiveness of the proposed framework on 128 univariate TSC datasets from the UCR archive\footnote{ \url{https://www.timeseriesclassification.com/}}, which are widely used for evaluating TSC methods~\cite{tang2020omni,ismail2020inceptiontime,wang2017time}. 
In these original datasets, the training and testing set have been well processed. We do not take any processing for these datasets for a fair comparison.

 \begin{table*}[h!]
    \setlength{\tabcolsep}{3pt}
    % {|>{\setlength{\tabcolsep}{3pt}}c|c|c|}
    \centering
    % \vspace{-0.2cm}
    \caption{Short-term TSF results of APTF for forecasting 1, 5, 8, and 10 future steps, where the average WMAPE over all horizons is reported. APTF also shows improvements over baselines CATS, TimeXer, and ModernTCN, as shown in Appendix Table~\ref{tab:main_res_short_term_new}.}
   \vspace{-0.1cm}
    \label{tab:main_res_shortterm}
    % \begin{tabular}{c|c|p{20pt}p{20pt}|cc|cc|cc|cc|cc}
    { \footnotesize
    \begin{tabular}{c|cc|cc|cc|cc|cc|cc|cc|cc}
        \hline
        \multirow{2}{*}{\shortstack{}} &  \multicolumn{2}{c|}{TimeMixer} &  \multicolumn{2}{c|}{PatchTST}&  \multicolumn{2}{c|}{Autoformer} &  \multicolumn{2}{c|}{NSformer} &  \multicolumn{2}{c|}{Scaleformer} &  \multicolumn{2}{c|}{NLinear} &  \multicolumn{2}{c|}{N-HiTS} &  \multicolumn{2}{c}{Informer} \\
         &  +APTF & Original & +APTF & Original & +APTF & Original & +APTF & Original & +APTF & Original & +APTF & Original& +APTF & Original  & +APTF & Original \\ 
        \midrule[0.5pt]
         \multirow{1}{*}{\rotatebox[origin=c]{0}{Fund1}} 
        &\textbf{87.914}&89.963&\textbf{86.549}&88.811&\textbf{103.286}&110.295&\textbf{95.112}&97.198&\textbf{100.027}&107.31&\textbf{102.536}&106.636&\textbf{86.856}&91.69&\textbf{102.912}&106.952 \\
        \midrule[0.5pt]
         \multirow{1}{*}{\rotatebox[origin=c]{0}{Fund2}} 
        &\textbf{84.356}&85.512&\textbf{85.094}&86.472&\textbf{99.712}&107.141&\textbf{91.516}&94.406&\textbf{96.736}&109.01&\textbf{90.791}&93.538&\textbf{86.13}&91.531&\textbf{91.407}&97.193 \\
                \midrule[0.5pt]
         \multirow{1}{*}{\rotatebox[origin=c]{0}{Fund3}} 
        &\textbf{83.659}&85.14&\textbf{84.821}&86.593&\textbf{98.838}&107.063&\textbf{90.364}&92.646&\textbf{95.235}&107.405&\textbf{90.542}&93.652&\textbf{84.719}&90.332&\textbf{90.476}&96.589 \\
 
        \hline

    \end{tabular}}
    % }
% \vspace{-0.1cm}
\end{table*}

\begin{table*}[h!]
    \setlength{\tabcolsep}{3pt}
    % {|>{\setlength{\tabcolsep}{3pt}}c|c|c|}
    \centering
    %% \vspace{-0.2cm}
    \caption{Long-term forecasting results for horizons of 96, 192, 336, and 720, reporting averaged MSE and MAE.
Due to space constraints, partial results for each horizon are provided in Appendix~\ref{tab:main_res_longterm_appendix}. APTF also shows improvements over baselines CATS, TimeXer, and ModernTCN, as shown in Appendix Table~\ref{tab:main_res_longterm_new} and Table~\ref{tab:full_longterm_p0}.}
   \vspace{-0.1cm}
    \label{tab:main_res_longterm}
    % \begin{tabular}{c|c|p{20pt}p{20pt}|cc|cc|cc|cc|cc}
    { \footnotesize
    \begin{tabular}{c|cc|cc|cc|cc|cc|cc|cc|cc}
        \hline
        \multirow{2}{*}{\shortstack{}}  &  \multicolumn{2}{c|}{TimeMixer} &  \multicolumn{2}{c|}{PatchTST}&  \multicolumn{2}{c|}{Autoformer} &  \multicolumn{2}{c|}{NSformer} &  \multicolumn{2}{c|}{Scaleformer} &  \multicolumn{2}{c|}{NLinear} &  \multicolumn{2}{c|}{N-HiTS} &  \multicolumn{2}{c}{Informer} \\
         &  +APTF & Original & +APTF & Original & +APTF & Original & +APTF & Original & +APTF & Original & +APTF & Original& +APTF & Original  & +APTF & Original \\ 
        \midrule[0.5pt]
         \multirow{1}{*}{\rotatebox[origin=c]{0}{ETTh1}} 
         &\textbf{0.492}&0.517&\textbf{0.40}&0.43&\textbf{0.492}&0.504&\textbf{0.511}&0.523&\textbf{0.543}&0.55&\textbf{0.453}&0.469&\textbf{0.491}&0.516&1.038&1.058\\
         
        \hline
         \multirow{1}{*}{\rotatebox[origin=c]{0}{ETTh2}} 
         &\textbf{0.395}&0.415&\textbf{0.366}&0.37&\textbf{0.398}&0.434&\textbf{0.407}&0.434&\textbf{0.409}&0.441&\textbf{0.38}&0.383&\textbf{0.546}&0.599&1.577&2.357 \\
 
       \hline
         \multirow{1}{*}{\rotatebox[origin=c]{0}{ETTm1}} 
        &\textbf{0.398}&0.407&\textbf{0.357}&0.369&\textbf{0.506}&0.556&\textbf{0.553}&0.599&\textbf{0.506}&0.518&\textbf{0.375}&0.377&\textbf{0.385}&0.399&0.675&0.74\\
         
        \hline
         \multirow{1}{*}{\rotatebox[origin=c]{0}{ETTm2}} 
        &\textbf{0.3}&0.304&\textbf{0.278}&0.29&\textbf{0.351}&0.36&\textbf{0.337}&0.356&\textbf{0.354}&0.414&\textbf{0.286}&0.291&\textbf{0.311}&0.334&1.55&2.092 \\
         
        \hline
         \multirow{1}{*}{\rotatebox[origin=c]{0}{weather}} 
        &\textbf{0.264}&0.269&\textbf{0.241}&0.244&\textbf{0.383}&0.386&\textbf{0.307}&0.314&\textbf{0.452}&0.534&\textbf{0.263}&0.27&\textbf{0.257}&0.262&0.508&0.595 \\
       \hline
         \multirow{1}{*}{\rotatebox[origin=c]{0}{Exchange}} 
         &\textbf{0.371}&0.381&\textbf{0.367}&0.374&\textbf{0.455}&0.488&\textbf{0.444}&0.46&\textbf{0.492}&0.562&\textbf{0.37}&0.376&\textbf{0.356}&0.402&0.875&0.924\\

      \hline
         \multirow{1}{*}{\rotatebox[origin=c]{0}{Electricity}} 
         &\textbf{0.288}&0.306&\textbf{0.25}&0.252&\textbf{0.335}&0.344&\textbf{0.321}&0.33&\textbf{0.336}&0.35&\textbf{0.319}&0.355&\textbf{0.255}&0.262&\textbf{0.737}&0.755\\

    \hline
         \multirow{1}{*}{\rotatebox[origin=c]{0}{Traffic}} 
          &\textbf{0.613}&0.648&\textbf{0.498}&0.501&\textbf{0.618}&0.642&\textbf{0.633}&0.661&\textbf{0.639}&0.665&\textbf{0.7}&0.796&\textbf{0.489}&0.503&\textbf{0.953}&0.976\\
         
        \hline
    \end{tabular}}
    % }
\vspace{-0.1cm}
\end{table*}

% \subsubsection{\textbf{Baselines.}}
\textbf{Baselines For TSF.} We use eleven advanced baselines to extensively evaluate our methods, including \textbf{one CNN-based} model, ModernTCN~\cite{luo2024moderntcn}, seven transformer-based models, CATS~\cite{kim2024self}, TimeXer~\cite{wangtimexer}, PatchTST~\cite{nie2022time_patchformer}, Autoformer~\cite{wu2021autoformer}, Non-stationary Transformer (NSformer)~\cite{liu2022non}, Scaleformer~\cite{shabani2022scaleformer}, and Informer~\cite{zhou2021informer}. 
We also include three linear models: TimeMixer, NLinear~\cite{zeng2023transformers_linear}, and N-HiTS~\cite{challu2022n_Nhits}. 
\textbf{For TSC, }five advanced baselines are selected for evaluating our training framework APTF, including InceptionTime~\cite{ismail2020inceptiontime}, OS-CNN~\cite{tang2020omni}, FCN, ResNet~\cite{wang2017time} and FormerTime~\cite{cheng2023formertime}. 

\textbf{Evaluation metrics.} Following the previous works, we use Mean Squared Error (MSE) and Mean Absolute Error (MAE) metrics for TSF methods evaluation~\cite{nie2022time_patchformer,wu2021autoformer,shabani2022scaleformer} while accuracy for TSC  evaluation~\cite{ismail2020inceptiontime,tang2020omni,cheng2023formertime}. 
Besides, the competition evaluation metric WMAPE (Weighted Mean Absolute Percentage Error) is also used to assess prediction accuracy, which measures the accuracy of predictions on larger transactions.

\textbf{Implementation details.}
When APTF is combined with baseline models, all experimental settings (e.g., model hyperparameters and optimizer) are consistent for fair comparison. More details are shown in Appendix~\ref{sec:Implementation} due to limited space.
For each experiment, we independently run four times with four different seeds, and the average metrics and standard deviations are reported. Experiments are conducted on NVIDIA GeForce RTX 3090 GPU on PyTorch. 

\textbf{Due to limitations in both the main text and the Appendix}, some results could not be fully presented, including (i) full results and standard deviations on each prediction length in short-term (corresponding to Table~\ref{tab:main_res_shortterm} and Table~\ref{tab:main_res_short_term_new}) and long-term TSF tasks (corresponding to Table~\ref{tab:main_res_longterm} and Table~\ref{tab:main_res_longterm_new}), and (ii) full results and standard deviations on each time series classification dataset (corresponding to Table~\ref{tab:metric_TSC}). 
We will show them in the full version paper.

\subsection{Results of time series forecasting (TSF) and time series classification (TSC)}

Low-predictability samples often yield large losses, causing training oscillations and poor local convergence.
\textbf{APTF mitigates this by maintaining normal learning on high-predictability samples while penalizing losses from low-predictability ones.}
This enables the model to learn useful patterns without overfitting noise, improving performance in both TSF and TSC tasks.
Due to space limitations, the results of short-term and long-term tasks for baselines CATS, TimeXer, and ModernTCN are presented in Appendix Table~\ref{tab:main_res_short_term_new} and Table~\ref{tab:main_res_longterm_new}.

\subsubsection{\textbf{Short-term forecasting}}
Results are presented in Table~\ref{tab:main_res_shortterm}. Across various prediction steps, our training framework APTF consistently improves the accuracy of eight baseline models, including five transformer-based models and three linear models.
Specifically, the average improvements for the transformer models are 2.06\%, 6.99\%, 2.56\%, 9.79\%, and 5.35\%, corresponding to models PatchTST, Autoformer, NSformer, Scaleformer, and Informer, respectively. 
The average improvements for the linear models are 1.79\%, 3.37\%, and 5.79\%, corresponding to TimeMixer, NLinear, and N-HiTS. 
The maximum improvements are more obvious.

\subsubsection{\textbf{Long-term forecasting}}
Results in Table~\ref{tab:main_res_longterm} also show that our training framework APTF consistently improves the accuracy of five transformer-based models and three linear models on different forecasting steps.
Specifically, the average improvements for the transformer models are 2.01\%, 4.78\%, 4.5\%, 8.15\%, and 13.14\%, corresponding to models PatchTST, Autoformer, NSformer, Scaleformer, and Informer, respectively. 
The average improvements for the linear models are 3.3\%, 3.24\%, and 5.75\%, corresponding to TimeMixer, NLinear, and N-HiTS. 
The maximum improvement ratios for them are more significant.

 \begin{table}[h]
 %% \vspace{-0.15cm}
    % \renewcommand{\arraystretch}{0.5}
    \setlength{\tabcolsep}{5.0pt}
    % {|>{\setlength{\tabcolsep}{3pt}}c|c|c|}
    \centering
    \caption{TSC task. The average classification accuracy over 128 UCR datasets is reported.}
   \vspace{-0.1cm}
    \label{tab:metric_TSC}
    % \begin{tabular}{c|c|p{20pt}p{20pt}|cc|cc|cc|cc|cc}
    {\footnotesize
    \begin{tabular}{c|c|cccc}
        \hline
        \multirow{2}{*}{\shortstack{}} &\multirow{2}{*}{Methods} &  \multicolumn{4}{c}{Average accuracy}  \\
         & & Origin & w/ APTF &Win &Tie \\ 
         \midrule[0.5pt]
         \multirow{5}{*}{\shortstack{128 UCR \\ datasets}} &FCNet &79.82$\pm$0.011&\textbf{81.21$\pm$0.014}&82&22\\
         
        &ResNet&79.93$\pm$0.0141&\textbf{81.04$\pm$0.0142}&86&25\\
        &InceptionTime &85.90$\pm$0.008&\textbf{86.73$\pm$0.008}&74&34\\

         &OSCNN &84.82$\pm$0.008&\textbf{85.37$\pm$0.0081}&65&40\\

    &FormerTime&74.42$\pm$0.0157&\textbf{75.31$\pm$0.0152}&69&33\\
    % \hline
    \hline
    &Average&80.97$\pm$0.0135&\textbf{81.93$\pm$0.0133}&75&31\\
        \hline
    \end{tabular}}
\vspace{-0.1cm}
\end{table}

\subsubsection{\textbf{Time series classification}}

% \subsection{Results of time series classification (TSC)}
Our method is not only effective in TSF task but also improves the accuracy of baseline models in TSC task, As shown in Table~\ref{tab:metric_TSC}. 
After applying our APTF, the baselines (FCNet, ResNet, InceptionTime, OSCNN, and FormerTime) achieved higher accuracy on an average of accuracy of 128 UCR datasets.
This indicates APTF is a general training framework for both TSF and TSC tasks.

 \begin{table*}[h!]
    \setlength{\tabcolsep}{1.25pt}
    % {|>{\setlength{\tabcolsep}{3pt}}c|c|c|}
    \centering
    %% \vspace{-0.2cm}
    \caption{Ablation study on fund sales datasets. The WMAPE of Fund1, Fund2, and Fund3 on each prediction step are averaged and reported. 
    ``w/o Amortize'' denotes not using the amortization model and ``w/o HPL'' denotes removing hierarchical buckets in Figure~\ref{fig:evolution_aware}(c) and the design of Figure~\ref{fig:evolution_aware}(b) is adopted. The best results are in bold.  }
   \vspace{-0.1cm}
    \label{tab:ablation}
    % \begin{tabular}{c|c|p{20pt}p{20pt}|cc|cc|cc|cc|cc}
    { \footnotesize
    \begin{tabular}{c|c|c|c|c||c|c|c||c|c|c||c|c|c}
        \hline
        \multirow{2}{*}{\shortstack{}} & &  \multicolumn{1}{c|}{\textbf{Autoformer-Our}} &  \multicolumn{1}{c|}{w/o Amortize}&  \multicolumn{1}{c||}{w/o HPL} &  \multicolumn{1}{c|}{\textbf{Scaleformer-Our}} &  \multicolumn{1}{c|}{w/o amortize}&  \multicolumn{1}{c||}{w/o HPL}&  \multicolumn{1}{c|}{\textbf{NSFormer-Our}} &  \multicolumn{1}{c|}{w/o Amortize}&  \multicolumn{1}{c||}{w/o HPL}&   \multicolumn{1}{c|}{\textbf{NLinear-Our}}&\multicolumn{1}{c|}{w/o Amortize}&  \multicolumn{1}{c}{w/o HPL} \\
      
        \midrule[0.5pt]
         \multirow{4}{*}{\rotatebox[origin=c]{90}{Fund-123}} 
        &1 &\textbf{87.15} &88.22 &88.29 &\textbf{90.63} &92.7 &93.19 &\textbf{84.99} &86.05 &86.22 &\textbf{89.44} &91.15 &92.55
        \\
        &5 &\textbf{104.42} &107.31 &108.33 &\textbf{98.12} &100.09 &103.04 &\textbf{94.94} &96.31 &96.96 &\textbf{92.11} &92.96 &94.37 \\
        &8 &\textbf{105.33} &106.13 &107.03 &\textbf{100.67} &102.26 &103.16 &\textbf{94.65} &96.84 &95.92 &\textbf{97.14} &97.67 &99.09 \\
        &10 &\textbf{106.31} &108.04 &107.88 &\textbf{101.29} &103.58 &102.3 &\textbf{95.77} &96.59 &96.38 &\textbf{99.8} &100.51 &101.62 \\
        \hline
    \end{tabular}}
    % }
\vspace{-0.1cm}
\end{table*}

\begin{table}[bt]
%% \vspace{-0.15cm}
    % \renewcommand{\arraystretch}{0.5}
    \setlength{\tabcolsep}{1pt}
    % {|>{\setlength{\tabcolsep}{3pt}}c|c|c|}
    \centering
    \caption{Further demonstrating the effectiveness of hierarchical predictability-aware loss (HPL) on two large-scale datasets. We report the standard deviations and average of MSE and MAE across 96, 192, 336, and 720 prediction steps.}
  \vspace{-0.1cm}
    \label{tab:compare_wo_HLP_further}
    % \begin{tabular}{c|c|p{20pt}p{20pt}|cc|cc|cc|cc|cc}
    % \small
    {\footnotesize
    \begin{tabular}{c|c|c|c|c|c|c|c}
        \hline
        \multirow{1}{*}{\shortstack{}} &  \multicolumn{1}{c|}{TimeMixer} &  \multicolumn{1}{c|}{PatchTST}&  \multicolumn{1}{c|}{Autoformer}  & \multicolumn{1}{c|}{NSFormer} & \multicolumn{1}{c|}{Scaleformer}& \multicolumn{1}{c|}{NLinear}& \multicolumn{1}{c}{NHits}\\
         % & & MSE  & MSE  & MSE  & MSE \\ 
         \midrule[0.5pt]
         \multirow{2}{*}{\shortstack{Electricity\\(w/o HPL)}}  &0.306&0.252&0.354&0.338&0.356&0.364&0.261\\
         &$\pm$4e-3&$\pm$3e-4&$\pm$0.01128&$\pm$0.01135&$\pm$0.017&$\pm$0.002&$\pm$2e-4\\
        \midrule[0.5pt]
        \multirow{2}{*}{\shortstack{Electricity\\(w/ HPL)}}  &\textbf{0.288}&\textbf{0.249}&\textbf{0.339}&\textbf{0.325}&\textbf{0.346}&\textbf{0.319}&\textbf{0.254}\\
        &$\pm$3e-3&$\pm$2e-4&$\pm$0.010&$\pm$0.007&$\pm$0.014&$\pm$0.001&$\pm$2e-4\\
         \midrule[0.5pt]
         \multirow{2}{*}{\shortstack{Traffic\\(w/o HPL)}}  &0.655&0.512&0.654&0.66&0.657&0.794&0.510\\
         &$\pm$0.013&$\pm$1e-3&$\pm$0.02&$\pm$0.012&$\pm$0.017&$\pm$3e-3&$\pm$6e-3\\
        \midrule[0.5pt]
        \multirow{2}{*}{\shortstack{Traffic\\(w/ HPL)}}  &\textbf{0.613}&\textbf{0.502}&\textbf{0.618}&\textbf{0.634}&\textbf{0.64}&\textbf{0.70}&\textbf{0.487}\\
        &$\pm$0.01&$\pm$1e-4&$\pm$0.018&$\pm$0.011&$\pm$0.02&$\pm$3e-3&$\pm$1e-3\\
        \midrule[0.5pt]

    \end{tabular}}
\vspace{-0.1cm}
\end{table}

\subsection{Ablation study}

\subsubsection{\textbf{Effectiveness of hierarchical predictability-aware loss (HPL)}}
HPL incorporates the previous bucket partitioning strategies into the loss calculation and averages the losses obtained from diverse bucket groups, As shown in Figure~\ref{fig:evolution_aware}(c) and algorithm~\ref{alg:algorithm2}.
This not only helps prevent large changes in gradient direction that can cause training oscillations but also compensates the gradients for highly predictable samples, avoiding the risk that they are not fully learned by the model. 

Here we remove the HPL design and use the predictability-aware loss with predictability evolution (Figure~\ref{fig:evolution_aware}(b)) for optimizing the model. 
The result of Table~\ref{tab:ablation} shows that using HPL can reduce forecasting errors by 2.8\%, 3.8\%, 1.7\%, and 2.9\% for baselines Autoformer, Scaleformer, NSformer, and NLinear. 

Moreover, \textbf{we further conduct an ablation study of HPL on two large-scale datasets electricity and traffic}, as shown in Table~\ref{tab:compare_wo_HLP_further}. 
Results indicate that HPL can significantly improve training performance, which further demonstrates the effectiveness of HPL.

\subsubsection{\textbf{Effectiveness of using amortization model}}
As shown in Table~\ref{tab:ablation}, compared to not using an amortization model, using it can reduce forecasting errors by 2.13\%, 2.11\%, 1.85\%, and 1.44\% for baselines Autoformer, Scaleformer, NSformer, and NLinear. 
This indicates that the amortization model can indeed help the source model amortize the predictability estimation errors caused by model bias, thereby further enhancing the performance of HPL.

\begin{figure}[bt]
%% \vspace{-0.3cm}
% \setlength{\abovecaptionskip}{0.1cm}
\centerline{\includegraphics[width=1\linewidth]{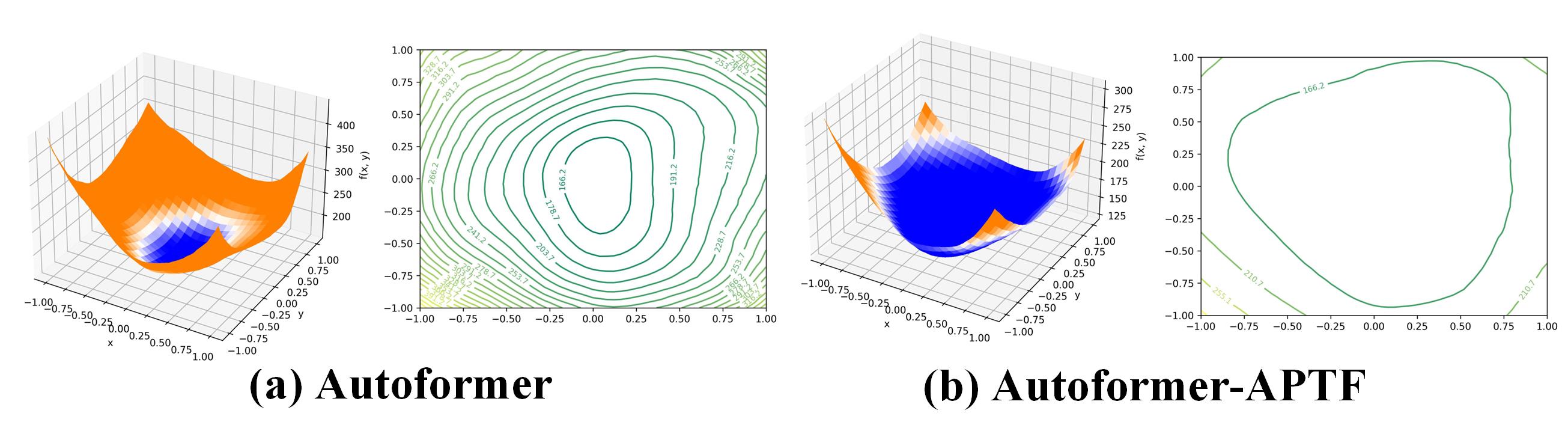} }
\vspace{-0.1cm}
\caption{Loss landscapes of Autoformer with and without APTF on the Fund sales dataset for 10-step forecasting.
A flatter minimum (larger blue area) indicates better generalization.
Additional cases are provided in Appendix Figure~\ref{fig:full_res_loss_landscape}.}
\label{fig:loss_land_scape}
\vspace{-0.1cm}
\end{figure}

\subsubsection{\textbf{Benefits on model generalization}}
It is known that the flatter the loss landscapes~\cite{DBLP:conf/nips/Li0TSG18} of the model, the better the robustness and generalization~\cite{DBLP:conf/iclr/ParkK22}. 
Figure~\ref{fig:loss_land_scape} shows that the loss landscape of Autoformer becomes significantly flatter after applying APTF, indicating that our framework can indeed improve the model's robustness and generalization by appropriately penalizing the low-predictability samples during training.

\begin{table}[h!]
%% \vspace{-0.15cm}
    % \renewcommand{\arraystretch}{0.5}
    \setlength{\tabcolsep}{3.25pt}
    % {|>{\setlength{\tabcolsep}{3pt}}c|c|c|}
    \centering
    \caption{Comparison with WaveBound and Co-teaching.
We report average performance across all prediction steps (average WMAPE for Fund and average MSE and MAE for other datasets).
Hyperparameter search results for Co-teaching are provided in Appendix Table~\ref{tab:hyper_coteaching}.}
   \vspace{-0.1cm}
    \label{tab:compare_wave}
    % \begin{tabular}{c|c|p{20pt}p{20pt}|cc|cc|cc|cc|cc}
    % \small
    {\footnotesize
    \begin{tabular}{c|c|c|c|c|c|c}
        \hline
        \multirow{1}{*}{\shortstack{}} &  \multicolumn{1}{c|}{ETTh1} &  \multicolumn{1}{c|}{ETTm1}&  \multicolumn{1}{c|}{weather}  & \multicolumn{1}{c|}{Fund1} & \multicolumn{1}{c|}{Fund2}& \multicolumn{1}{c}{Fund3}\\
         % & & MSE  & MSE  & MSE  & MSE \\ 
         \midrule[0.5pt]
         \multirow{1}{*}{\shortstack{TimeMixer-Original}}  &0.517&0.404&0.269&\underline{89.963}&\underline{85.512}&\underline{85.14}\\
        \midrule[0.5pt]
        % \midrule[0.5pt]
        \multirow{1}{*}{\shortstack{ -Wavebound}}  &0.706&0.489&0.279&206.9&186.198&328.27\\
        \midrule[0.5pt]
        \multirow{1}{*}{\shortstack{ -Coteaching}}  &\underline{0.508}&\underline{0.402}&\underline{0.268}&100.23&95.716&94.893\\
        \midrule[0.5pt]
        \multirow{1}{*}{\shortstack{ -APTF (Ours)}}  &\textbf{0.492}&\textbf{0.396}&\textbf{0.264}&\textbf{87.91}&\textbf{84.36}&\textbf{83.659}\\
        \midrule[0.5pt]
        % \hline
        \multirow{1}{*}{\shortstack{PatchTST-Original}}  &0.430&0.369&\underline{0.244}&88.811&\underline{86.472}&\underline{86.593}\\
        \midrule[0.5pt]
        % \midrule[0.5pt]
        \multirow{1}{*}{\shortstack{ -Wavebound}}  &0.434&\underline{0.364}&0.245&191.085&197.374&201.886\\
        \midrule[0.5pt]
        \multirow{1}{*}{\shortstack{ -Coteaching}}  &\underline{0.429}&0.37&0.245&\underline{88.604}&88.668&87.022\\
        \midrule[0.5pt]
        \multirow{1}{*}{\shortstack{ -APTF (Ours)}}  &\textbf{0.420}&\textbf{0.357}&\textbf{0.242}&\textbf{86.55}&\textbf{85.09}&\textbf{84.821}\\
        \midrule[0.5pt]
        
        \multirow{1}{*}{\shortstack{NSFormer-Original}} &0.520&0.599&0.312&97.199&\underline{94.406}&\underline{92.646}\\
        \midrule[0.5pt]
        % \midrule[0.5pt]
        \multirow{1}{*}{\shortstack{ -Wavebound}}  &0.528&0.595&\underline{0.309}&193.569&181.116&211.739\\
        \midrule[0.5pt]
        \multirow{1}{*}{\shortstack{ -Coteaching}}  &\underline{0.516}&\underline{0.582}&0.312&\underline{96.744}&94.847&93.348\\
        \midrule[0.5pt]
        \multirow{1}{*}{\shortstack{ -APTF (Ours)}}  &\textbf{0.511}&\textbf{0.553}&\textbf{0.307}&\textbf{95.364}&\textbf{91.826}&\textbf{90.579}\\
        \midrule[0.5pt]
        
        % \midrule[0.5pt]
    \end{tabular}}
\vspace{-0.1cm}
\end{table} 
\subsubsection{\textbf{Compared with framework Co-teaching and Wavebound}}
The comparison results between APTF and baseline training frameworks Wavebound~\cite{cho2022wavebound} and Co-teaching~\cite{han2018co} are shown in Table~\ref{tab:compare_wave}.
The results show that APTF outperforms both of them, especially for the fund sales dataset that suffers from high noise due to complex factors such as unexpected policy and public opinion. 
The advantage of our method lies in imposing appropriate penalties on the losses of samples, mitigating the adverse effects caused by excessively large losses of low-predictability samples, and enabling the model to better learn from both highly predictable and low-predictability samples. 
In contrast, Co-teaching's approach of directly discarding samples with high loss values may lead to knowledge waste and harm model performance. 
Additionally, when the error falls below a certain threshold, Wavebound collectively applies gradient ascent to all samples, which may impact the model's ability to learn from some highly predictable samples.

\begin{figure*}[bt]
%% \vspace{-0.3cm}
% \setlength{\abovecaptionskip}{0.1cm}
\centerline{\includegraphics[width=1\linewidth]{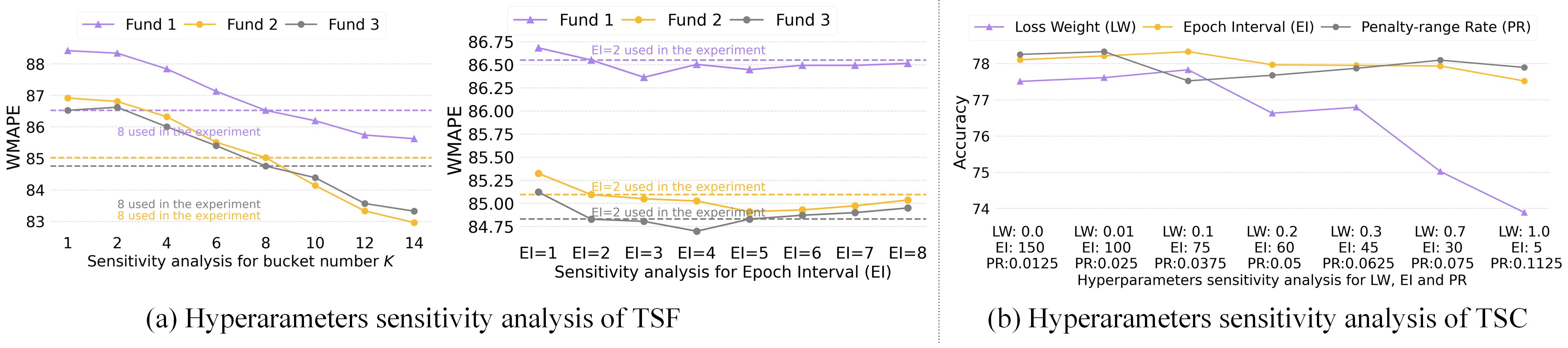} }
\vspace{-0.1cm}
\caption{Hyperparameter sensitivity analysis.}
\label{fig:hyper_sensitvity}
\vspace{-0.1cm}
\end{figure*}

\subsubsection{\textbf{Compared with predictability-aware loss functions}}
Table~\ref{tab:compare_loss_base} shows that our APTF achieves better overall performance than adaptive robust loss (ARL) and Self-paced learning (SPL). In addition, we also empirically observe that APTF significantly outperforms ARL and SPL on three Fund datasets that suffer from high noise, further demonstrating its effectiveness. 
Overall, the experiment results demonstrate that our APTF can better address low-predictability samples to improve model performance.
In contrast, SPL directly discards low-predictability samples in the early stages, similar to Co-teaching, which may lead to underfitting as the model fails to learn sufficiently from the knowledge contained in low-predictability samples. 
The poor performance of ARL indicates that relying solely on adaptive parameters may not impose sufficient constraints on low-predictability samples, thereby impairing model convergence.

\begin{table}[h!]
%% \vspace{-0.15cm}
    % \renewcommand{\arraystretch}{0.5}
    \setlength{\tabcolsep}{3pt}
    % {|>{\setlength{\tabcolsep}{3pt}}c|c|c|}
    \centering
    \caption{Comparison with predictability-aware loss functions on the Electricity dataset.
We report the average MSE and MAE over prediction horizons of 96, 192, 336, and 720 steps.
ARL~\cite{barron2019general} denotes adaptive robust loss, while SPL~\cite{kumar2010self} denotes self-paced learning loss.}
   \vspace{-0.1cm}
    \label{tab:compare_loss_base}
    % \begin{tabular}{c|c|p{20pt}p{20pt}|cc|cc|cc|cc|cc}
    % \small
    {\footnotesize
    \begin{tabular}{c|c|c|c|c|c|c}
        \hline
        \multirow{1}{*}{\shortstack{}} &  \multicolumn{1}{c|}{TimeMixer} &  \multicolumn{1}{c|}{Autoformer}  & \multicolumn{1}{c|}{NSFormer} & \multicolumn{1}{c|}{Scaleformer}& \multicolumn{1}{c|}{NLinear}& \multicolumn{1}{c}{NHits}\\
         % & & MSE  & MSE  & MSE  & MSE \\ 
         \midrule[0.5pt]
        \multirow{1}{*}{\shortstack{Original}}  &0.306&0.344&0.330&0.35&0.355&0.262\\
         \multirow{1}{*}{\shortstack{+APTF}}  &\textbf{0.288}&\textbf{0.335}&\textbf{0.321}&\textbf{0.336}&\textbf{0.319}&\textbf{0.255}\\
        \multirow{1}{*}{\shortstack{+ARL}}  &0.307&0.357&0.329&0.347&0.367&0.273\\
        \multirow{1}{*}{\shortstack{+SPL}}  &0.300&0.349&0.332&0.451&0.360&0.262\\

        \midrule[0.5pt]

    \end{tabular}}
\vspace{-0.1cm}
\end{table}

\subsection{Time complexity and memory cost analysis}
When using the amortized model, our method has approximately double the memory overhead and time complexity of the original model. 
Without amortizing, there are almost no additional costs. 
For instance, on the weather dataset, the average training epoch time is \textbf{51.75s} \textit{VS} \textbf{52.99s} for PatchTST \textit{VS} Ours (PatchTST-APTF-w/o amortization), \textbf{52.89s} \textit{VS} \textbf{53.20s} for Scaleformer \textit{VS} Ours, and \textbf{40.07s} \textit{VS} \textbf{42.75s} for Autoformer \textit{VS} Ours.
More importantly, our method still brings significant accuracy improvements to the baseline model without using the amortized model, as shown in Tables~\ref{tab:main_res_longterm} and Table~\ref{tab:compare_wo_amortize}. 
Therefore, \textbf{our method can improve model performance without obviously increasing model overhead when not using the amortization model}, which further demonstrates the effectiveness of our approach.
\begin{table}[h!]
%% \vspace{-0.15cm}
    % \renewcommand{\arraystretch}{0.5}
    \setlength{\tabcolsep}{2.5pt}
    % {|>{\setlength{\tabcolsep}{3pt}}c|c|c|}
    \centering
    \caption{APTF can still improve baseline accuracy \textbf{without introducing additional time complexity or memory overhead} when the amortization model is not used.
We report the average WMAPE for the Fund datasets and the average MSE for the other datasets.}
   \vspace{-0.1cm}
    \label{tab:compare_wo_amortize}
    % \begin{tabular}{c|c|p{20pt}p{20pt}|cc|cc|cc|cc|cc}
    % \small
    {\footnotesize
    \begin{tabular}{c|c|c|c|c|c|c|c}
        \hline
        \multirow{1}{*}{\shortstack{}} &  \multicolumn{1}{c|}{ET.h1} &  \multicolumn{1}{c|}{ET.h2}&  \multicolumn{1}{c|}{ET.m1}  & \multicolumn{1}{c|}{ET.m2} & \multicolumn{1}{c|}{Fund1}& \multicolumn{1}{c|}{Fund2}& \multicolumn{1}{c}{Fund3}\\
         % & & MSE  & MSE  & MSE  & MSE \\ 
         \midrule[0.5pt]
         \multirow{1}{*}{\shortstack{TimeMixer-Original}}  &0.537&0.405&0.402&0.281&89.96&85.51&85.14\\
        \midrule[0.5pt]
        \multirow{1}{*}{\shortstack{-Ours (w/o amortization)}}  &\textbf{0.512}&\textbf{0.385}&\textbf{0.396}&\textbf{0.279}&\textbf{88.5}&\textbf{84.62}&\textbf{84.02}\\
        \midrule[0.5pt]
        % \hline
        \multirow{1}{*}{\shortstack{PatchTST-Original}}  &0.423&0.347&0.351&0.258&88.81&86.47&86.59\\
        \midrule[0.5pt]
        % \midrule[0.5pt]
        \multirow{1}{*}{\shortstack{-Ours (w/o amortization)}}  &\textbf{0.416}&\textbf{0.345}&\textbf{0.345}&\textbf{0.250}&\textbf{87.2}&\textbf{85.7}&\textbf{85.45}\\
        \midrule[0.5pt]

        \multirow{1}{*}{\shortstack{NHiTS-Original}} &0.526&0.612&0.386&0.307&91.69&91.53&90.33\\
        \midrule[0.5pt]
        % \midrule[0.5pt]
        \multirow{1}{*}{\shortstack{-Ours (w/o amortization)}}  &\textbf{0.499}&\textbf{0.582}&\textbf{0.373}&\textbf{0.291}&\textbf{87.9}&\textbf{87.1}&\textbf{85.8}\\
        \midrule[0.5pt]
        
        \multirow{1}{*}{\shortstack{NSFormer-Original}} &0.539&0.419&0.657&0.335&97.2&94.41&92.65\\
        \midrule[0.5pt]
        % \midrule[0.5pt]
        \multirow{1}{*}{\shortstack{-Ours (w/o amortization)}}  &\textbf{0.535}&\textbf{0.396}&\textbf{0.618}&\textbf{0.318}&\textbf{95.97}&\textbf{92.34}&\textbf{91.73}\\
        \midrule[0.5pt]
        
        % \midrule[0.5pt]
    \end{tabular}}
\vspace{-0.1cm}
\end{table} 

\subsection{Hyperparameter sensitivity analysis}
We analyze hyperparameters such as the number of buckets and epoch intervals.
As shown in Figure~\ref{fig:hyper_sensitvity}, for TSF, we observe that the prediction error continuously decreases with the increasing number of buckets in each hierarchical bucket group. 
This is reasonable because more buckets result in a finer bucket division and the low-predictability samples will be penalized more accurately, thereby improving model performance. 
When the number of buckets exceeds 15, the improvement in model performance tends to saturate. It is worth noting that the maximum number of buckets in the experiments is set to 9, which is not optimal. 
We also observe that the better epoch interval $\varepsilon$ is 3 or 4, while used in the paper is 2, which is not optimal. 
For TSC, we observe that the loss weight, epoch interval, and penalty range rate are set as 0.1, 75, and 0.025, respectively can achieve overall good performance.

\section{Conclusion}

This paper proposes the Amortized Predictability-aware Training Framework (APTF) to enhance the training of baseline models in both TSF and TSC tasks. APTF boosts model convergence through two key designs: the Hierarchical Predictability-aware Loss (HPL) and the amortization model. These components enable the model to focus on high-predictability samples while still learning appropriately from low-predictability ones, with a reduced risk of overfitting to noise data, thereby improving overall training performance. Extensive experiments on popular TSF and TSC datasets demonstrate that APTF consistently enhances the convergence of diverse advanced models, including eleven TSF and five TSC baselines.

% This paper proposes the Amortized Predictability-aware Training Framework (APTF) to improve the training effect of baseline models in both TSF and TSC tasks. APTF improves the model convergence based on the key designs of Hierarchical Predictability-aware Loss (HPL) and the amortization model.
% The two designs enable the TS model to dynamically find the low-predictability samples and reduce their gradient contributions, mitigating the risk of overfitting noise in them, thereby improving training performance.
% Extensive experiments on popular TSF datasets and TSC datasets show that APTF can improve the convergence of various advanced models, including eight TSF and five TSC baselines.

\section{Limitations}
Despite its generality and effectiveness, APTF still has room for improvement in terms of hyperparameter configuration.
Key hyperparameters in APTF, such as the number of buckets $K$ and the stage interval $\varepsilon$, currently require manual specification. $K$ controls the granularity of predictability-based sample partitioning, where a larger $K$ enables finer distinction but may reduce per-bucket reliability under limited data. $\varepsilon$ governs how quickly the penalization range expands to cover more low-predictability samples, where a smaller $\varepsilon$ yields faster noise suppression but risks training instability. Since optimal settings may vary across datasets and training dynamics, adaptively determining and dynamically adjusting these hyperparameters based on dataset characteristics and time series patterns remains a promising direction for future work.

\section{Acknowledgements}
This work was supported by the National Natural Science Foundation of China under Grant 62427819.


\bibliographystyle{ACM-Reference-Format}
\balance
\bibliography{sample-base}


\begin{thebibliography}{46}

%%% the Web address.

\ifx \showCODEN    \undefined \def \showCODEN     #1{\unskip}     \fi
\ifx \showISBNx    \undefined \def \showISBNx     #1{\unskip}     \fi
\ifx \showISBNxiii \undefined \def \showISBNxiii  #1{\unskip}     \fi
\ifx \showISSN     \undefined \def \showISSN      #1{\unskip}     \fi
\ifx \showLCCN     \undefined \def \showLCCN      #1{\unskip}     \fi
\ifx \shownote     \undefined \def \shownote      #1{#1}          \fi
\ifx \showarticletitle \undefined \def \showarticletitle #1{#1}   \fi
\ifx \showURL      \undefined \def \showURL       {\relax}        \fi
% invisible to TeX
\providecommand\bibfield[2]{#2}
\providecommand\bibinfo[2]{#2}
\providecommand\natexlab[1]{#1}
\providecommand\showeprint[2][]{arXiv:#2}

\bibitem[Arpit et~al\mbox{.}(2017)]%
        {arpit2017closer}
\bibfield{author}{\bibinfo{person}{Devansh Arpit}, \bibinfo{person}{Stanis{\l}aw Jastrz{\k{e}}bski}, \bibinfo{person}{Nicolas Ballas}, \bibinfo{person}{David Krueger}, \bibinfo{person}{Emmanuel Bengio}, \bibinfo{person}{Maxinder~S Kanwal}, \bibinfo{person}{Tegan Maharaj}, \bibinfo{person}{Asja Fischer}, \bibinfo{person}{Aaron Courville}, \bibinfo{person}{Yoshua Bengio}, {et~al\mbox{.}}} \bibinfo{year}{2017}\natexlab{}.
\newblock \showarticletitle{A closer look at memorization in deep networks}. In \bibinfo{booktitle}{\emph{International conference on machine learning}}. PMLR, \bibinfo{pages}{233--242}.
\newblock


\bibitem[Balcan et~al\mbox{.}(2006)]%
        {balcan2006agnostic}
\bibfield{author}{\bibinfo{person}{Maria-Florina Balcan}, \bibinfo{person}{Alina Beygelzimer}, {and} \bibinfo{person}{John Langford}.} \bibinfo{year}{2006}\natexlab{}.
\newblock \showarticletitle{Agnostic active learning}. In \bibinfo{booktitle}{\emph{Proceedings of the 23rd international conference on Machine learning}}. \bibinfo{pages}{65--72}.
\newblock


\bibitem[Barron(2019)]%
        {barron2019general}
\bibfield{author}{\bibinfo{person}{Jonathan~T Barron}.} \bibinfo{year}{2019}\natexlab{}.
\newblock \showarticletitle{A general and adaptive robust loss function}. In \bibinfo{booktitle}{\emph{Proceedings of the IEEE/CVF conference on computer vision and pattern recognition}}. \bibinfo{pages}{4331--4339}.
\newblock


\bibitem[Blum and Mitchell(1998)]%
        {blum1998combining}
\bibfield{author}{\bibinfo{person}{Avrim Blum} {and} \bibinfo{person}{Tom Mitchell}.} \bibinfo{year}{1998}\natexlab{}.
\newblock \showarticletitle{Combining labeled and unlabeled data with co-training}. In \bibinfo{booktitle}{\emph{Proceedings of the eleventh annual conference on Computational learning theory}}. \bibinfo{pages}{92--100}.
\newblock


\bibitem[Challu et~al\mbox{.}(2022)]%
        {challu2022n_Nhits}
\bibfield{author}{\bibinfo{person}{C Challu}, \bibinfo{person}{KG Olivares}, \bibinfo{person}{BN Oreshkin}, \bibinfo{person}{F Garza}, \bibinfo{person}{M Mergenthaler}, {and} \bibinfo{person}{A Dubrawski}.} \bibinfo{year}{2022}\natexlab{}.
\newblock \showarticletitle{N-hits: Neural hierarchical interpolation for time series forecasting. arXiv}.
\newblock \bibinfo{journal}{\emph{arXiv preprint arXiv:2201.12886}} (\bibinfo{year}{2022}).
\newblock


\bibitem[Chen et~al\mbox{.}(2023)]%
        {chen2023tsmixer}
\bibfield{author}{\bibinfo{person}{Si-An Chen}, \bibinfo{person}{Chun-Liang Li}, \bibinfo{person}{Nate Yoder}, \bibinfo{person}{Sercan~O Arik}, {and} \bibinfo{person}{Tomas Pfister}.} \bibinfo{year}{2023}\natexlab{}.
\newblock \showarticletitle{Tsmixer: An all-mlp architecture for time series forecasting}.
\newblock \bibinfo{journal}{\emph{arXiv preprint arXiv:2303.06053}} (\bibinfo{year}{2023}).
\newblock


\bibitem[Cheng et~al\mbox{.}(2023)]%
        {cheng2023formertime}
\bibfield{author}{\bibinfo{person}{Mingyue Cheng}, \bibinfo{person}{Qi Liu}, \bibinfo{person}{Zhiding Liu}, \bibinfo{person}{Zhi Li}, \bibinfo{person}{Yucong Luo}, {and} \bibinfo{person}{Enhong Chen}.} \bibinfo{year}{2023}\natexlab{}.
\newblock \showarticletitle{Formertime: Hierarchical multi-scale representations for multivariate time series classification}. In \bibinfo{booktitle}{\emph{Proceedings of the ACM Web Conference 2023}}. \bibinfo{pages}{1437--1445}.
\newblock


\bibitem[Cho et~al\mbox{.}(2014)]%
        {cho2014learning}
\bibfield{author}{\bibinfo{person}{Kyunghyun Cho}, \bibinfo{person}{Bart Van~Merri{\"e}nboer}, \bibinfo{person}{Caglar Gulcehre}, \bibinfo{person}{Dzmitry Bahdanau}, \bibinfo{person}{Fethi Bougares}, \bibinfo{person}{Holger Schwenk}, {and} \bibinfo{person}{Yoshua Bengio}.} \bibinfo{year}{2014}\natexlab{}.
\newblock \showarticletitle{Learning phrase representations using RNN encoder-decoder for statistical machine translation}.
\newblock \bibinfo{journal}{\emph{arXiv preprint arXiv:1406.1078}} (\bibinfo{year}{2014}).
\newblock


\bibitem[Cho et~al\mbox{.}(2022)]%
        {cho2022wavebound}
\bibfield{author}{\bibinfo{person}{Youngin Cho}, \bibinfo{person}{Daejin Kim}, \bibinfo{person}{Dongmin Kim}, \bibinfo{person}{Mohammad~Azam Khan}, {and} \bibinfo{person}{Jaegul Choo}.} \bibinfo{year}{2022}\natexlab{}.
\newblock \showarticletitle{WaveBound: dynamic error bounds for stable time series forecasting}.
\newblock \bibinfo{journal}{\emph{Advances in Neural Information Processing Systems}}  \bibinfo{volume}{35} (\bibinfo{year}{2022}), \bibinfo{pages}{19579--19591}.
\newblock


\bibitem[Das et~al\mbox{.}(2023)]%
        {DBLP:journals/tmlr/DasKLMSY23}
\bibfield{author}{\bibinfo{person}{Abhimanyu Das}, \bibinfo{person}{Weihao Kong}, \bibinfo{person}{Andrew Leach}, \bibinfo{person}{Shaan Mathur}, \bibinfo{person}{Rajat Sen}, {and} \bibinfo{person}{Rose Yu}.} \bibinfo{year}{2023}\natexlab{}.
\newblock \showarticletitle{Long-term Forecasting with TiDE: Time-series Dense Encoder}.
\newblock \bibinfo{journal}{\emph{Trans. Mach. Learn. Res.}}  \bibinfo{volume}{2023} (\bibinfo{year}{2023}).
\newblock
\urldef\tempurl%
\url{https://openreview.net/forum?id=pCbC3aQB5W}
\showURL{%
\tempurl}


\bibitem[Diebold and Strasser(2013)]%
        {diebold2013correlation}
\bibfield{author}{\bibinfo{person}{Francis~X Diebold} {and} \bibinfo{person}{Georg Strasser}.} \bibinfo{year}{2013}\natexlab{}.
\newblock \showarticletitle{On the correlation structure of microstructure noise: A financial economic approach}.
\newblock \bibinfo{journal}{\emph{Review of Economic Studies}} \bibinfo{volume}{80}, \bibinfo{number}{4} (\bibinfo{year}{2013}), \bibinfo{pages}{1304--1337}.
\newblock


\bibitem[Forestier et~al\mbox{.}(2018)]%
        {forestier2018surgical}
\bibfield{author}{\bibinfo{person}{Germain Forestier}, \bibinfo{person}{Fran{\c{c}}ois Petitjean}, \bibinfo{person}{Pavel Senin}, \bibinfo{person}{Fabien Despinoy}, \bibinfo{person}{Arnaud Huaulm{\'e}}, \bibinfo{person}{Hassan~Ismail Fawaz}, \bibinfo{person}{Jonathan Weber}, \bibinfo{person}{Lhassane Idoumghar}, \bibinfo{person}{Pierre-Alain Muller}, {and} \bibinfo{person}{Pierre Jannin}.} \bibinfo{year}{2018}\natexlab{}.
\newblock \showarticletitle{Surgical motion analysis using discriminative interpretable patterns}.
\newblock \bibinfo{journal}{\emph{Artificial intelligence in medicine}}  \bibinfo{volume}{91} (\bibinfo{year}{2018}), \bibinfo{pages}{3--11}.
\newblock


\bibitem[Freund et~al\mbox{.}(1999)]%
        {freund1999short}
\bibfield{author}{\bibinfo{person}{Yoav Freund}, \bibinfo{person}{Robert Schapire}, {and} \bibinfo{person}{Naoki Abe}.} \bibinfo{year}{1999}\natexlab{}.
\newblock \showarticletitle{A short introduction to boosting}.
\newblock \bibinfo{journal}{\emph{Journal-Japanese Society For Artificial Intelligence}} \bibinfo{volume}{14}, \bibinfo{number}{771-780} (\bibinfo{year}{1999}), \bibinfo{pages}{1612}.
\newblock


\bibitem[Han et~al\mbox{.}(2018)]%
        {han2018co}
\bibfield{author}{\bibinfo{person}{Bo Han}, \bibinfo{person}{Quanming Yao}, \bibinfo{person}{Xingrui Yu}, \bibinfo{person}{Gang Niu}, \bibinfo{person}{Miao Xu}, \bibinfo{person}{Weihua Hu}, \bibinfo{person}{Ivor Tsang}, {and} \bibinfo{person}{Masashi Sugiyama}.} \bibinfo{year}{2018}\natexlab{}.
\newblock \showarticletitle{Co-teaching: Robust training of deep neural networks with extremely noisy labels}.
\newblock \bibinfo{journal}{\emph{Advances in neural information processing systems}}  \bibinfo{volume}{31} (\bibinfo{year}{2018}).
\newblock


\bibitem[Huang et~al\mbox{.}(2022)]%
        {huang2022semi}
\bibfield{author}{\bibinfo{person}{Tao Huang}, \bibinfo{person}{Pengfei Chen}, {and} \bibinfo{person}{Ruipeng Li}.} \bibinfo{year}{2022}\natexlab{}.
\newblock \showarticletitle{A semi-supervised vae based active anomaly detection framework in multivariate time series for online systems}. In \bibinfo{booktitle}{\emph{Proceedings of the ACM Web Conference 2022}}. \bibinfo{pages}{1797--1806}.
\newblock


\bibitem[Ismail~Fawaz et~al\mbox{.}(2019)]%
        {Ismail_Muller_2019}
\bibfield{author}{\bibinfo{person}{Hassan Ismail~Fawaz}, \bibinfo{person}{Germain Forestier}, \bibinfo{person}{Jonathan Weber}, \bibinfo{person}{Lhassane Idoumghar}, {and} \bibinfo{person}{Pierre-Alain Muller}.} \bibinfo{year}{2019}\natexlab{}.
\newblock \showarticletitle{Deep learning for time series classification: a review}.
\newblock \bibinfo{journal}{\emph{Data Mining and Knowledge Discovery}} (\bibinfo{date}{Jul} \bibinfo{year}{2019}), \bibinfo{pages}{917–963}.
\newblock
\href{https://doi.org/10.1007/s10618-019-00619-1}{doi:\nolinkurl{10.1007/s10618-019-00619-1}}


\bibitem[Ismail~Fawaz et~al\mbox{.}(2020)]%
        {ismail2020inceptiontime}
\bibfield{author}{\bibinfo{person}{Hassan Ismail~Fawaz}, \bibinfo{person}{Benjamin Lucas}, \bibinfo{person}{Germain Forestier}, \bibinfo{person}{Charlotte Pelletier}, \bibinfo{person}{Daniel~F Schmidt}, \bibinfo{person}{Jonathan Weber}, \bibinfo{person}{Geoffrey~I Webb}, \bibinfo{person}{Lhassane Idoumghar}, \bibinfo{person}{Pierre-Alain Muller}, {and} \bibinfo{person}{Fran{\c{c}}ois Petitjean}.} \bibinfo{year}{2020}\natexlab{}.
\newblock \showarticletitle{Inceptiontime: Finding alexnet for time series classification}.
\newblock \bibinfo{journal}{\emph{Data Mining and Knowledge Discovery}} \bibinfo{volume}{34}, \bibinfo{number}{6} (\bibinfo{year}{2020}), \bibinfo{pages}{1936--1962}.
\newblock


\bibitem[Jiang et~al\mbox{.}(2018)]%
        {jiang2018mentornet}
\bibfield{author}{\bibinfo{person}{Lu Jiang}, \bibinfo{person}{Zhengyuan Zhou}, \bibinfo{person}{Thomas Leung}, \bibinfo{person}{Li-Jia Li}, {and} \bibinfo{person}{Li Fei-Fei}.} \bibinfo{year}{2018}\natexlab{}.
\newblock \showarticletitle{Mentornet: Learning data-driven curriculum for very deep neural networks on corrupted labels}. In \bibinfo{booktitle}{\emph{International conference on machine learning}}. PMLR, \bibinfo{pages}{2304--2313}.
\newblock


\bibitem[Jordan(1997)]%
        {jordan1997serial}
\bibfield{author}{\bibinfo{person}{Michael~I Jordan}.} \bibinfo{year}{1997}\natexlab{}.
\newblock \showarticletitle{Serial order: A parallel distributed processing approach}.
\newblock In \bibinfo{booktitle}{\emph{Advances in psychology}}. Vol.~\bibinfo{volume}{121}. \bibinfo{publisher}{Elsevier}, \bibinfo{pages}{471--495}.
\newblock


\bibitem[Kim et~al\mbox{.}(2024)]%
        {kim2024self}
\bibfield{author}{\bibinfo{person}{D Kim}, \bibinfo{person}{J Park}, \bibinfo{person}{J Lee}, {and} \bibinfo{person}{H Kim}.} \bibinfo{year}{2024}\natexlab{}.
\newblock \showarticletitle{Are Self-Attentions Effective for Time Series Forecasting?}. In \bibinfo{booktitle}{\emph{38th Conference on Neural Information Processing Systems (NeurIPS 2024)}}.
\newblock


\bibitem[Kumar et~al\mbox{.}(2010)]%
        {kumar2010self}
\bibfield{author}{\bibinfo{person}{M Kumar}, \bibinfo{person}{Benjamin Packer}, {and} \bibinfo{person}{Daphne Koller}.} \bibinfo{year}{2010}\natexlab{}.
\newblock \showarticletitle{Self-paced learning for latent variable models}.
\newblock \bibinfo{journal}{\emph{Advances in neural information processing systems}}  \bibinfo{volume}{23} (\bibinfo{year}{2010}).
\newblock


\bibitem[Li et~al\mbox{.}(2018)]%
        {DBLP:conf/nips/Li0TSG18}
\bibfield{author}{\bibinfo{person}{Hao Li}, \bibinfo{person}{Zheng Xu}, \bibinfo{person}{Gavin Taylor}, \bibinfo{person}{Christoph Studer}, {and} \bibinfo{person}{Tom Goldstein}.} \bibinfo{year}{2018}\natexlab{}.
\newblock \showarticletitle{Visualizing the loss landscape of neural nets}.
\newblock \bibinfo{journal}{\emph{Advances in neural information processing systems}}  \bibinfo{volume}{31} (\bibinfo{year}{2018}).
\newblock


\bibitem[Liu et~al\mbox{.}(2022)]%
        {liu2022non}
\bibfield{author}{\bibinfo{person}{Yong Liu}, \bibinfo{person}{Haixu Wu}, \bibinfo{person}{Jianmin Wang}, {and} \bibinfo{person}{Mingsheng Long}.} \bibinfo{year}{2022}\natexlab{}.
\newblock \showarticletitle{Non-stationary transformers: Exploring the stationarity in time series forecasting}.
\newblock \bibinfo{journal}{\emph{Advances in Neural Information Processing Systems}}  \bibinfo{volume}{35} (\bibinfo{year}{2022}), \bibinfo{pages}{9881--9893}.
\newblock


\bibitem[Luo and Wang(2024)]%
        {luo2024moderntcn}
\bibfield{author}{\bibinfo{person}{Donghao Luo} {and} \bibinfo{person}{Xue Wang}.} \bibinfo{year}{2024}\natexlab{}.
\newblock \showarticletitle{Moderntcn: A modern pure convolution structure for general time series analysis}. In \bibinfo{booktitle}{\emph{The twelfth international conference on learning representations}}. \bibinfo{pages}{1--43}.
\newblock


\bibitem[Nie et~al\mbox{.}({[n.\,d.]})]%
        {nie2022time_patchformer}
\bibfield{author}{\bibinfo{person}{Yuqi Nie}, \bibinfo{person}{Nam~H Nguyen}, \bibinfo{person}{Phanwadee Sinthong}, {and} \bibinfo{person}{Jayant Kalagnanam}.} \bibinfo{year}{[n.\,d.]}\natexlab{}.
\newblock \showarticletitle{A Time Series is Worth 64 Words: Long-term Forecasting with Transformers}. In \bibinfo{booktitle}{\emph{The Eleventh International Conference on Learning Representations}}.
\newblock


\bibitem[Nunes et~al\mbox{.}(2023)]%
        {nunes2023challenges}
\bibfield{author}{\bibinfo{person}{P Nunes}, \bibinfo{person}{J Santos}, {and} \bibinfo{person}{E Rocha}.} \bibinfo{year}{2023}\natexlab{}.
\newblock \showarticletitle{Challenges in predictive maintenance--A review}.
\newblock \bibinfo{journal}{\emph{CIRP Journal of Manufacturing Science and Technology}}  \bibinfo{volume}{40} (\bibinfo{year}{2023}), \bibinfo{pages}{53--67}.
\newblock


\bibitem[Pandit et~al\mbox{.}(2017)]%
        {pandit2017noise}
\bibfield{author}{\bibinfo{person}{Diptangshu Pandit}, \bibinfo{person}{Li Zhang}, \bibinfo{person}{Chengyu Liu}, \bibinfo{person}{Nauman Aslam}, \bibinfo{person}{Samiran Chattopadhyay}, {and} \bibinfo{person}{Chee~Peng Lim}.} \bibinfo{year}{2017}\natexlab{}.
\newblock \showarticletitle{Noise reduction in ECG signals using wavelet transform and dynamic thresholding}.
\newblock \bibinfo{journal}{\emph{Emerging trends in neuro engineering and neural computation}} (\bibinfo{year}{2017}), \bibinfo{pages}{193--206}.
\newblock


\bibitem[Park and Kim(2022)]%
        {DBLP:conf/iclr/ParkK22}
\bibfield{author}{\bibinfo{person}{Namuk Park} {and} \bibinfo{person}{Songkuk Kim}.} \bibinfo{year}{2022}\natexlab{}.
\newblock \showarticletitle{How Do Vision Transformers Work?}. In \bibinfo{booktitle}{\emph{The Tenth International Conference on Learning Representations, {ICLR} 2022, Virtual Event, April 25-29, 2022}}. \bibinfo{publisher}{OpenReview.net}.
\newblock
\urldef\tempurl%
\url{https://openreview.net/forum?id=D78Go4hVcxO}
\showURL{%
\tempurl}


\bibitem[Sarkar and Etemad(2020)]%
        {sarkar2020self}
\bibfield{author}{\bibinfo{person}{Pritam Sarkar} {and} \bibinfo{person}{Ali Etemad}.} \bibinfo{year}{2020}\natexlab{}.
\newblock \showarticletitle{Self-supervised ECG representation learning for emotion recognition}.
\newblock \bibinfo{journal}{\emph{IEEE Transactions on Affective Computing}} \bibinfo{volume}{13}, \bibinfo{number}{3} (\bibinfo{year}{2020}), \bibinfo{pages}{1541--1554}.
\newblock


\bibitem[Sen et~al\mbox{.}(2019)]%
        {sen2019think}
\bibfield{author}{\bibinfo{person}{Rajat Sen}, \bibinfo{person}{Hsiang-Fu Yu}, {and} \bibinfo{person}{Inderjit~S Dhillon}.} \bibinfo{year}{2019}\natexlab{}.
\newblock \showarticletitle{Think globally, act locally: A deep neural network approach to high-dimensional time series forecasting}.
\newblock \bibinfo{journal}{\emph{Advances in neural information processing systems}}  \bibinfo{volume}{32} (\bibinfo{year}{2019}).
\newblock


\bibitem[Shabani et~al\mbox{.}(2022)]%
        {shabani2022scaleformer}
\bibfield{author}{\bibinfo{person}{Amin Shabani}, \bibinfo{person}{Amir Abdi}, \bibinfo{person}{Lili Meng}, {and} \bibinfo{person}{Tristan Sylvain}.} \bibinfo{year}{2022}\natexlab{}.
\newblock \showarticletitle{Scaleformer: iterative multi-scale refining transformers for time series forecasting}.
\newblock \bibinfo{journal}{\emph{arXiv preprint arXiv:2206.04038}} (\bibinfo{year}{2022}).
\newblock


\bibitem[Tang et~al\mbox{.}(2022)]%
        {tang2020omni}
\bibfield{author}{\bibinfo{person}{Wensi Tang}, \bibinfo{person}{Guodong Long}, \bibinfo{person}{Lu Liu}, \bibinfo{person}{Tianyi Zhou}, \bibinfo{person}{Michael Blumenstein}, {and} \bibinfo{person}{Jing Jiang}.} \bibinfo{year}{2022}\natexlab{}.
\newblock \showarticletitle{Omni-Scale CNNs: a simple and effective kernel size configuration for time series classification}. In \bibinfo{booktitle}{\emph{The Tenth International Conference on Learning Representations, {ICLR} 2022, Virtual Event, April 25-29, 2022}}. \bibinfo{publisher}{OpenReview.net}.
\newblock
\urldef\tempurl%
\url{https://openreview.net/forum?id=PDYs7Z2XFGv}
\showURL{%
\tempurl}


\bibitem[Tang et~al\mbox{.}(2020)]%
        {tang2020rethinking}
\bibfield{author}{\bibinfo{person}{Wensi Tang}, \bibinfo{person}{Guodong Long}, \bibinfo{person}{Lu Liu}, \bibinfo{person}{Tianyi Zhou}, \bibinfo{person}{Jing Jiang}, {and} \bibinfo{person}{Michael Blumenstein}.} \bibinfo{year}{2020}\natexlab{}.
\newblock \showarticletitle{Rethinking 1d-cnn for time series classification: A stronger baseline}.
\newblock \bibinfo{journal}{\emph{arXiv preprint arXiv:2002.10061}} (\bibinfo{year}{2020}), \bibinfo{pages}{1--7}.
\newblock


\bibitem[Vaswani et~al\mbox{.}(2017)]%
        {vaswani2017attention}
\bibfield{author}{\bibinfo{person}{Ashish Vaswani}, \bibinfo{person}{Noam Shazeer}, \bibinfo{person}{Niki Parmar}, \bibinfo{person}{Jakob Uszkoreit}, \bibinfo{person}{Llion Jones}, \bibinfo{person}{Aidan~N Gomez}, \bibinfo{person}{{\L}ukasz Kaiser}, {and} \bibinfo{person}{Illia Polosukhin}.} \bibinfo{year}{2017}\natexlab{}.
\newblock \showarticletitle{Attention is all you need}.
\newblock \bibinfo{journal}{\emph{Advances in neural information processing systems}}  \bibinfo{volume}{30} (\bibinfo{year}{2017}).
\newblock


\bibitem[Wang et~al\mbox{.}(2024)]%
        {wangtimemixer}
\bibfield{author}{\bibinfo{person}{Shiyu Wang}, \bibinfo{person}{Haixu Wu}, \bibinfo{person}{Xiaoming Shi}, \bibinfo{person}{Tengge Hu}, \bibinfo{person}{Huakun Luo}, \bibinfo{person}{Lintao Ma}, \bibinfo{person}{James~Y Zhang}, {and} \bibinfo{person}{Jun Zhou}.} \bibinfo{year}{2024}\natexlab{}.
\newblock \showarticletitle{TimeMixer: Decomposable Multiscale Mixing for Time Series Forecasting}. In \bibinfo{booktitle}{\emph{The Twelfth International Conference on Learning Representations}}.
\newblock


\bibitem[Wang et~al\mbox{.}({[n.\,d.]})]%
        {wangtimexer}
\bibfield{author}{\bibinfo{person}{Yuxuan Wang}, \bibinfo{person}{Haixu Wu}, \bibinfo{person}{Jiaxiang Dong}, \bibinfo{person}{Guo Qin}, \bibinfo{person}{Haoran Zhang}, \bibinfo{person}{Yong Liu}, \bibinfo{person}{Yunzhong Qiu}, \bibinfo{person}{Jianmin Wang}, {and} \bibinfo{person}{Mingsheng Long}.} \bibinfo{year}{[n.\,d.]}\natexlab{}.
\newblock \showarticletitle{TimeXer: Empowering Transformers for Time Series Forecasting with Exogenous Variables}. In \bibinfo{booktitle}{\emph{The Thirty-eighth Annual Conference on Neural Information Processing Systems}}.
\newblock


\bibitem[Wang et~al\mbox{.}(2017)]%
        {wang2017time}
\bibfield{author}{\bibinfo{person}{Zhiguang Wang}, \bibinfo{person}{Weizhong Yan}, {and} \bibinfo{person}{Tim Oates}.} \bibinfo{year}{2017}\natexlab{}.
\newblock \showarticletitle{Time series classification from scratch with deep neural networks: A strong baseline}. In \bibinfo{booktitle}{\emph{2017 International joint conference on neural networks (IJCNN)}}. IEEE, \bibinfo{pages}{1578--1585}.
\newblock


\bibitem[Wu et~al\mbox{.}(2021)]%
        {wu2021autoformer}
\bibfield{author}{\bibinfo{person}{Haixu Wu}, \bibinfo{person}{Jiehui Xu}, \bibinfo{person}{Jianmin Wang}, {and} \bibinfo{person}{Mingsheng Long}.} \bibinfo{year}{2021}\natexlab{}.
\newblock \showarticletitle{Autoformer: Decomposition transformers with auto-correlation for long-term series forecasting}.
\newblock \bibinfo{journal}{\emph{Advances in Neural Information Processing Systems}}  \bibinfo{volume}{34} (\bibinfo{year}{2021}), \bibinfo{pages}{22419--22430}.
\newblock


\bibitem[Zeng et~al\mbox{.}(2023)]%
        {zeng2023transformers_linear}
\bibfield{author}{\bibinfo{person}{Ailing Zeng}, \bibinfo{person}{Muxi Chen}, \bibinfo{person}{Lei Zhang}, {and} \bibinfo{person}{Qiang Xu}.} \bibinfo{year}{2023}\natexlab{}.
\newblock \showarticletitle{Are transformers effective for time series forecasting?}. In \bibinfo{booktitle}{\emph{Proceedings of the AAAI conference on artificial intelligence}}, Vol.~\bibinfo{volume}{37}. \bibinfo{pages}{11121--11128}.
\newblock


\bibitem[Zhang et~al\mbox{.}(2021)]%
        {zhang2021understanding}
\bibfield{author}{\bibinfo{person}{Chiyuan Zhang}, \bibinfo{person}{Samy Bengio}, \bibinfo{person}{Moritz Hardt}, \bibinfo{person}{Benjamin Recht}, {and} \bibinfo{person}{Oriol Vinyals}.} \bibinfo{year}{2021}\natexlab{}.
\newblock \showarticletitle{Understanding deep learning (still) requires rethinking generalization}.
\newblock \bibinfo{journal}{\emph{Commun. ACM}} \bibinfo{volume}{64}, \bibinfo{number}{3} (\bibinfo{year}{2021}), \bibinfo{pages}{107--115}.
\newblock


\bibitem[Zhang et~al\mbox{.}(2024)]%
        {zhang2024self}
\bibfield{author}{\bibinfo{person}{Xu Zhang}, \bibinfo{person}{Zhengang Huang}, \bibinfo{person}{Yunzhi Wu}, \bibinfo{person}{Xun Lu}, \bibinfo{person}{Erpeng Qi}, \bibinfo{person}{Yunkai Chen}, \bibinfo{person}{Zhongya Xue}, \bibinfo{person}{Peng Wang}, {and} \bibinfo{person}{Wei Wang}.} \bibinfo{year}{2024}\natexlab{}.
\newblock \showarticletitle{Self-Adaptive Scale Handling for Forecasting Time Series with Scale Heterogeneity}. In \bibinfo{booktitle}{\emph{ICASSP 2024-2024 IEEE International Conference on Acoustics, Speech and Signal Processing (ICASSP)}}. IEEE, \bibinfo{pages}{7485--7489}.
\newblock


\bibitem[Zhang et~al\mbox{.}(2025a)]%
        {zhang2025multi}
\bibfield{author}{\bibinfo{person}{Xu Zhang}, \bibinfo{person}{Zhengang Huang}, \bibinfo{person}{Yunzhi Wu}, \bibinfo{person}{Xun Lu}, \bibinfo{person}{Erpeng Qi}, \bibinfo{person}{Yunkai Chen}, \bibinfo{person}{Zhongya Xue}, \bibinfo{person}{Qitong Wang}, \bibinfo{person}{Peng Wang}, {and} \bibinfo{person}{Wei Wang}.} \bibinfo{year}{2025}\natexlab{a}.
\newblock \showarticletitle{Multi-period Learning for Financial Time Series Forecasting}. In \bibinfo{booktitle}{\emph{Proceedings of the 31st ACM SIGKDD Conference on Knowledge Discovery and Data Mining V. 1}}. \bibinfo{pages}{2848--2859}.
\newblock


\bibitem[Zhang et~al\mbox{.}(2025c)]%
        {zhang2025global}
\bibfield{author}{\bibinfo{person}{Xu Zhang}, \bibinfo{person}{Peng Wang}, \bibinfo{person}{Chen Wang}, \bibinfo{person}{Zhe Xu}, \bibinfo{person}{Xiaohua Nie}, {and} \bibinfo{person}{Wei Wang}.} \bibinfo{year}{2025}\natexlab{c}.
\newblock \showarticletitle{Global Feature Enhancing and Fusion Framework for Strain Gauge Status Recognition}. In \bibinfo{booktitle}{\emph{Companion Proceedings of the ACM on Web Conference 2025}}. \bibinfo{pages}{611--620}.
\newblock


\bibitem[Zhang et~al\mbox{.}(2025b)]%
        {zhang2025lightweight}
\bibfield{author}{\bibinfo{person}{Xu Zhang}, \bibinfo{person}{Qitong Wang}, \bibinfo{person}{Peng Wang}, {and} \bibinfo{person}{Wei Wang}.} \bibinfo{year}{2025}\natexlab{b}.
\newblock \showarticletitle{A Lightweight Sparse Interaction Network for Time Series Forecasting}. In \bibinfo{booktitle}{\emph{Proceedings of the AAAI Conference on Artificial Intelligence}}, Vol.~\bibinfo{volume}{39}. \bibinfo{pages}{13304--13312}.
\newblock


\bibitem[Zhou et~al\mbox{.}(2021)]%
        {zhou2021informer}
\bibfield{author}{\bibinfo{person}{Haoyi Zhou}, \bibinfo{person}{Shanghang Zhang}, \bibinfo{person}{Jieqi Peng}, \bibinfo{person}{Shuai Zhang}, \bibinfo{person}{Jianxin Li}, \bibinfo{person}{Hui Xiong}, {and} \bibinfo{person}{Wancai Zhang}.} \bibinfo{year}{2021}\natexlab{}.
\newblock \showarticletitle{Informer: Beyond efficient transformer for long sequence time-series forecasting}. In \bibinfo{booktitle}{\emph{Proceedings of the AAAI conference on artificial intelligence}}, Vol.~\bibinfo{volume}{35}. \bibinfo{pages}{11106--11115}.
\newblock


\bibitem[Zhou et~al\mbox{.}(2022)]%
        {zhou2022film}
\bibfield{author}{\bibinfo{person}{Tian Zhou}, \bibinfo{person}{Ziqing Ma}, \bibinfo{person}{Qingsong Wen}, \bibinfo{person}{Liang Sun}, \bibinfo{person}{Tao Yao}, \bibinfo{person}{Wotao Yin}, \bibinfo{person}{Rong Jin}, {et~al\mbox{.}}} \bibinfo{year}{2022}\natexlab{}.
\newblock \showarticletitle{Film: Frequency improved legendre memory model for long-term time series forecasting}.
\newblock \bibinfo{journal}{\emph{Advances in Neural Information Processing Systems}}  \bibinfo{volume}{35} (\bibinfo{year}{2022}), \bibinfo{pages}{12677--12690}.
\newblock


\end{thebibliography}

\appendix

\section{Appendices}
\subsection{Experimental settings}
\subsubsection{\textbf{Implementation details}}
\label{sec:Implementation}
Time Series Forecasting (TSF). Training uses a fixed epoch interval $\varepsilon=2$ and 30 epochs for all methods, with input lengths of 96 or 512. The initial number of buckets is 9 and decreases by 1 at each training stage. Loss weights in the first stage are uniformly decreasing: the first bucket has weight 1, the last is half of the previous bucket, and intermediate buckets decrease evenly with interval $1/(K-1)$, where $K$ is the number of buckets. In subsequent stages, the largest weight is removed.
Time Series Classification (TSC). Each batch is split into high- and low-predictability buckets with loss weights 1 and $\in{0.1, 0.01}$, respectively. Training runs for 300 epochs with $\varepsilon=75$, gradually increasing the proportion of low-predictability samples by 0.025 per stage to expand gradient penalties.

\begin{figure}[h!]
% \vspace{-0.3cm}
\centerline{\includegraphics[width=1\linewidth]{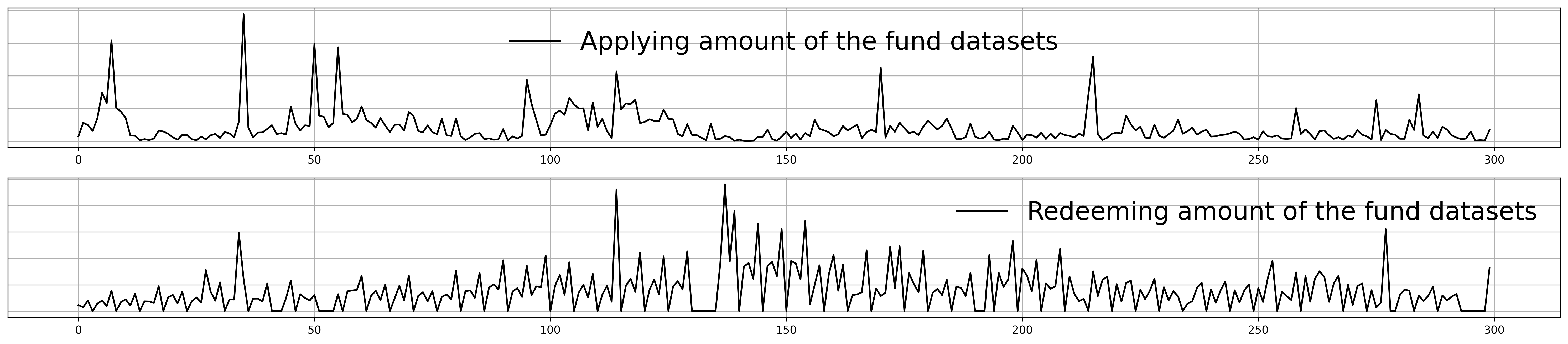} }
\vspace{-0.2cm} %调整的是Figure字样和图片的距离
\caption{Time series visualization of Fund sales dataset.}
\label{fig:series_comp_appen}
\vspace{-0.4cm}
\end{figure}

\subsubsection{\textbf{Discussion and visualization of Fund sales datasets (Figure~\ref{fig:series_comp_appen})}}
\label{sec:fund_discuss}
Fund sales time series exhibit sharp trends and frequent fluctuations due to market conditions, policies, and public opinion (Figure~\ref{fig:series_comp_appen}), leading to many low-predictability samples that hinder training. This makes the fund dataset well suited for evaluating our framework’s ability to identify and penalize such samples, where improved accuracy indicates better handling of unpredictability and enhanced convergence.

\subsection{Additional experiment results}

\subsubsection{\textbf{ More baselines (CATS, ModernTCN and TimeXer) on short-term forecasting (Table~\ref{tab:main_res_short_term_new}) and long-term forecasting (Table~\ref{tab:main_res_longterm_new})}}
Here, we further demonstrate the application of APTF to CATS, ModernTCN, and TimeXer for both short-term and long-term forecasting. The results in Table~\ref{tab:main_res_short_term_new} and Table~\ref{tab:main_res_longterm_new}) show that APTF consistently and significantly improves the training performance of these baselines, further proving that APTF is generalizable, universal, and a model-agnostic framework.

\subsubsection{\textbf{Hyperparameter search of Co-teaching for fair comparison (Table~\ref{tab:hyper_coteaching})}}
As shown in Table~\ref{tab:hyper_coteaching}, we have conducted a hyperparameter search for Co-teaching, and its best performance is still inferior to ours, compared with Table~\ref{tab:compare_wave} of the manuscript. 
This indicates that the approach of directly discarding high-loss samples in Co-teaching indeed has limitations, while our method effectively addresses this issue and achieves better performance.

\begin{figure*}[h]
\centerline{\includegraphics[width=\linewidth]{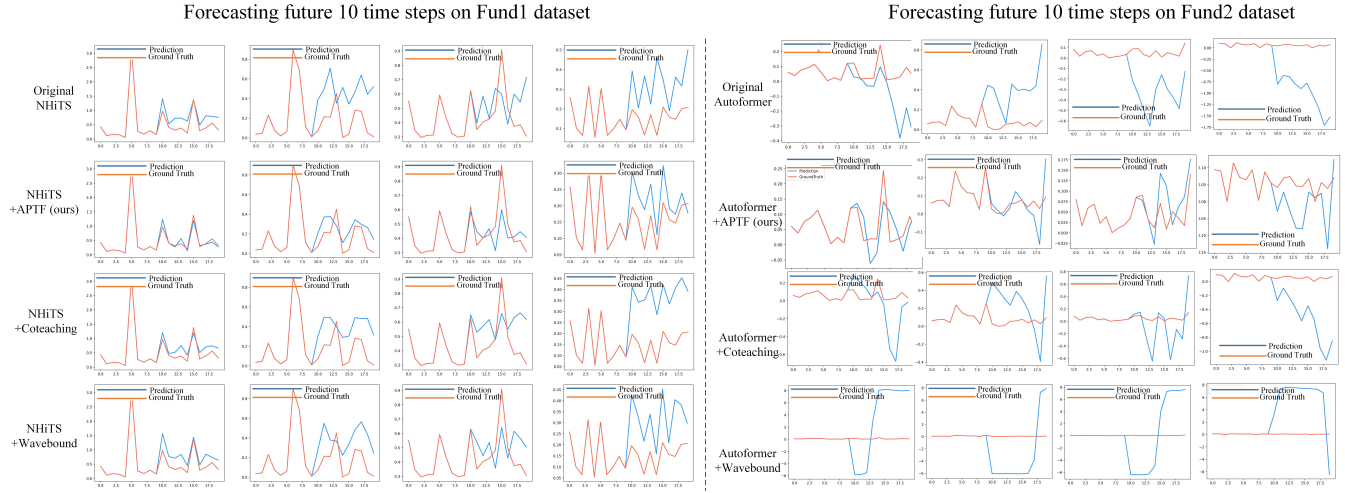} }
\caption{Visualizations of forecasting future 10-time steps on Fund1 and Fund2 datasets with the original baseline NHiTS (left) and Autoformer (right), as well as the versions of applying APTF (ours), Coteaching, and Wavebound.}
\label{fig:model}
%% \vspace{-0.4cm}
\label{fig:case_vis_fund_1}
\end{figure*}

\subsubsection{\textbf{Visualization of forecasting (Figure~\ref{fig:case_vis_fund_1})}}
We further provide qualitative visualizations to illustrate APTF’s effectiveness. In Figure~\ref{fig:case_vis_fund_1}, models trained with APTF better capture overall trends, indicating that dynamically penalizing low-predictability samples and emphasizing high-predictability ones improves performance.

\begin{figure*}[h]
\centerline{\includegraphics[width=\linewidth]{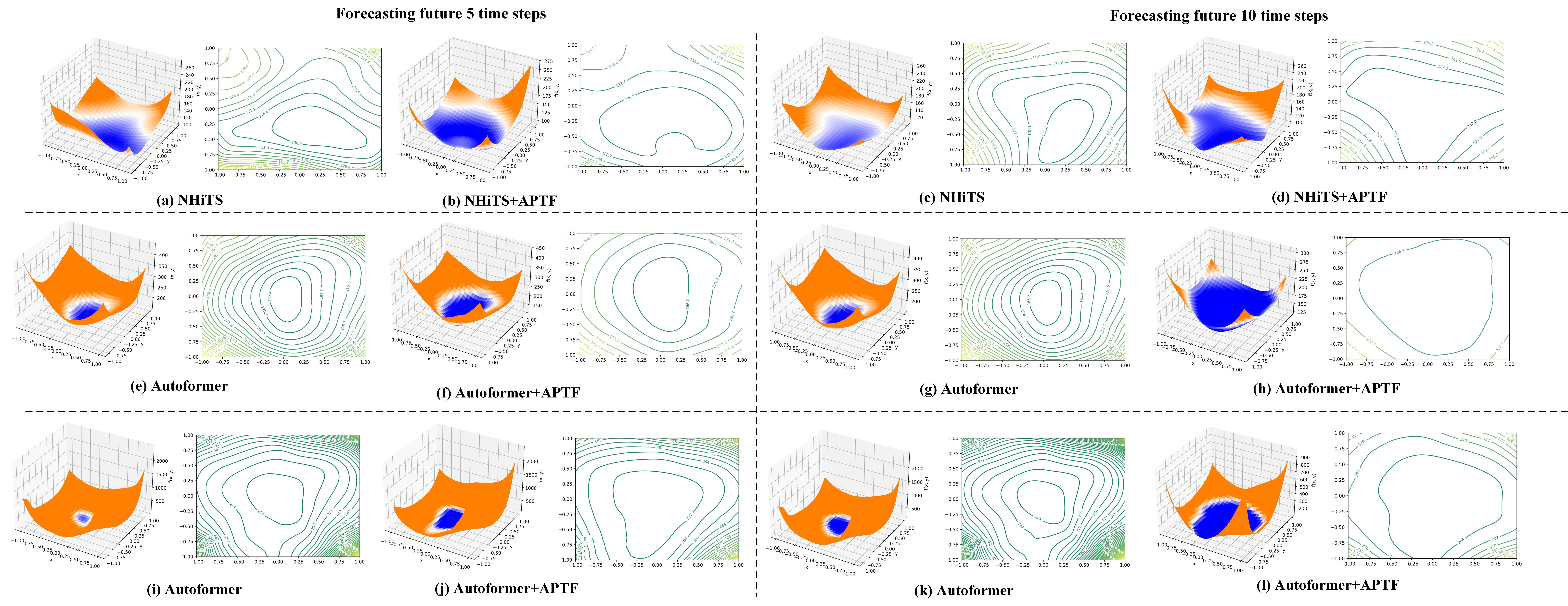} }
\caption{Full results of loss landscapes on different baselines with the Fund1 dataset. The flatter the bottom (the larger the blue area) indicates the better the generalization of the trained models.}
\label{fig:model}
%% \vspace{-0.4cm}
\label{fig:full_res_loss_landscape}
\end{figure*}

\subsubsection{\textbf{Full results on model generalization (Figure~\ref{fig:full_res_loss_landscape})}}
We also present more experimental results on whether APTF can bring generalizability to models, as shown in Figure~\ref{fig:full_res_loss_landscape}. The results indicate that training based on the APTF framework tends to make models converge to flatter regions of the loss landscape, thereby achieving better generalizability.

\subsubsection{\textbf{Results of long-term forecasting on each prediction length (Table~\ref{tab:full_longterm_p0} and Table~\ref{tab:main_res_longterm_appendix})}}
The results show that APTF consistently improves the performance of baseline models across different prediction lengths, which illustrates the effectiveness and generalizability of the APTF framework.

\begin{table}[h]
    \setlength{\tabcolsep}{4.3pt}
    % {|>{\setlength{\tabcolsep}{3pt}}c|c|c|}
    \centering
    %% \vspace{-0.2cm}
    \caption{Additional baselines for short-term TSF with average WMAPE on 1, 5, 8, and 10 future steps. Results for each prediction length are omitted due to limited space.}
   \vspace{-0.2cm}
    \label{tab:main_res_short_term_new}
    % \begin{tabular}{c|c|p{20pt}p{20pt}|cc|cc|cc|cc|cc}
    { \small
    \begin{tabular}{c|cc|cc|cc}
        \hline
        \multirow{2}{*}{\shortstack{}}  &  \multicolumn{2}{c|}{CATS} &  \multicolumn{2}{c|}{ModernTCN}&  \multicolumn{2}{c}{TimeXer}  \\
         &  +APTF & Original & +APTF & Original & +APTF & Original  \\ 
        \midrule[0.5pt]
         \multirow{1}{*}{\rotatebox[origin=c]{0}{Fund1}} 
 &\textbf{85.72} &88.17&\textbf{86.47} &90.28&\textbf{83.97} &86.29\\
                  \multirow{1}{*}{\rotatebox[origin=c]{0}{Fund2}} &\textbf{82.92} &84.79&\textbf{83.83} &85.07&\textbf{82.1} &83.55\\

                  \multirow{1}{*}{\rotatebox[origin=c]{0}{Fund3}} &\textbf{80.56} &83.91&\textbf{82.55} &85.16&\textbf{81.38} &84.02\\

        \hline
    \end{tabular}}
    % }
\vspace{-0.3cm}
\end{table}

\begin{table}[h]
    \setlength{\tabcolsep}{3.5pt}
    % {|>{\setlength{\tabcolsep}{3pt}}c|c|c|}
    \centering
    %% \vspace{-0.2cm}
    \caption{Additional baselines for long-term TSF with average MSE and MAE on future 96, 192, 336, and 720 steps. Table~\ref{tab:full_longterm_p0} shows results for each prediction length.}
   \vspace{-0.2cm}
    \label{tab:main_res_longterm_new}
    % \begin{tabular}{c|c|p{20pt}p{20pt}|cc|cc|cc|cc|cc}
    { \small
    \begin{tabular}{c|cc|cc|cc}
        \hline
        \multirow{2}{*}{\shortstack{}}  &  \multicolumn{2}{c|}{CATS} &  \multicolumn{2}{c|}{ModernTCN}&  \multicolumn{2}{c}{TimeXer}  \\
         &  +APTF & Original & +APTF & Original & +APTF & Original  \\ 
        \midrule[0.5pt]
         \multirow{1}{*}{\rotatebox[origin=c]{0}{ETTh1}} 
         &\textbf{0.426} &0.430&\textbf{0.426} &0.438&\textbf{0.436} &0.445\\
                  \multirow{1}{*}{\rotatebox[origin=c]{0}{ETTh2}} 
&\textbf{0.363} &0.373&\textbf{0.365} &0.374&\textbf{0.367} &0.373\\
                  \multirow{1}{*}{\rotatebox[origin=c]{0}{ETTm1}} 
&\textbf{0.357} &0.366&\textbf{0.371} &0.386&\textbf{0.377} &0.382\\
                  \multirow{1}{*}{\rotatebox[origin=c]{0}{ETTm2}} 
 &\textbf{0.288} &0.313&\textbf{0.289} &0.302&\textbf{0.284} &0.295\\
                  \multirow{1}{*}{\rotatebox[origin=c]{0}{Weather}} 
&\textbf{0.245} &0.246&\textbf{0.246} &0.251&\textbf{0.248} &0.249\\
                  \multirow{1}{*}{\rotatebox[origin=c]{0}{Exchange}} 
&\textbf{0.364} &0.376&\textbf{0.369} &0.38&\textbf{0.378} &0.388\\

                           \multirow{1}{*}{\rotatebox[origin=c]{0}{Electricity}} 
&\textbf{0.270} &0.275&\textbf{0.235} &0.243&\textbf{0.284} &0.289\\
                           \multirow{1}{*}{\rotatebox[origin=c]{0}{Traffic }} 
 &\textbf{0.611} &0.627&\textbf{0.509} &0.529&\textbf{0.558} &0.564\\

        \hline
    \end{tabular}}
    % }
\vspace{-0.5cm}
\end{table}

\begin{table}[h]
%% \vspace{-0.15cm}
    % \renewcommand{\arraystretch}{0.5}
    \setlength{\tabcolsep}{2.25pt}
    % {|>{\setlength{\tabcolsep}{3pt}}c|c|c|}
    \centering
    \caption{Hyperparameter search for Coteaching. We report the average of MSE and MAE on predicting all timesteps. The forget rate $r$ determines the number of samples discarded during the training process.}
   % \vspace{-0.2cm}
    \label{tab:hyper_coteaching}
    % \begin{tabular}{c|c|p{20pt}p{20pt}|cc|cc|cc|cc|cc}
    \small{
    \begin{tabular}{c|c|c|c|c|c|c}
        \hline
        \multirow{1}{*}{\shortstack{Forget rate $r$}} &  \multicolumn{1}{c|}{ETTh1} &  \multicolumn{1}{c|}{ETTm1}&  \multicolumn{1}{c|}{Weather}  & \multicolumn{1}{c|}{Fund1} & \multicolumn{1}{c|}{Fund2}
        & \multicolumn{1}{c}{Fund3}\\
         % & & MSE  & MSE  & MSE  & MSE \\ 

        \midrule[0.5pt]
        \multirow{1}{*}{TimeMixer, \shortstack{\textit{r}=0.1}} &0.509&0.403&0.268&100.23&95.716&94.893\\

         \midrule[0.5pt]
        \multirow{1}{*}{TimeMixer, \shortstack{\textit{r}=0.2}} &0.508&0.403&0.269&101.29&96.11&95.571\\

       \midrule[0.5pt]
        \multirow{1}{*}{TimeMixer, \shortstack{\textit{r}=0.3}}  &0.508&0.402&0.27&102.278&96.846&96.057\\
          \midrule[0.5pt]
        \multirow{1}{*}{TimeMixer, \shortstack{\textit{r}=0.4}}   &0.509&0.402&0.27&102.972&97.133&96.714\\
      \midrule[0.5pt]
      \midrule[0.5pt]
    %%%%%%%%%%%%%%%%%%%
        \multirow{1}{*}{PatchTST, \shortstack{\textit{r}=0.1}} &0.429&0.37&0.245&88.604&88.668&87.022\\

         \midrule[0.5pt]
        \multirow{1}{*}{PatchTST, \shortstack{\textit{r}=0.2}} &0.429&0.371&0.246&89.706&90.582&88.074\\

       \midrule[0.5pt]
        \multirow{1}{*}{PatchTST, \shortstack{\textit{r}=0.3}}  &0.429&0.372&0.246&91.421&94.287&89.612\\
          \midrule[0.5pt]
        \multirow{1}{*}{PatchTST, \shortstack{\textit{r}=0.4}}  &0.43&0.372&0.246&92.81&97.306&92.284\\
      \midrule[0.5pt]
          \midrule[0.5pt]
    %%%%%%%%%%%%%%%%%%%
        \multirow{1}{*}{NSFormer, \shortstack{\textit{r}=0.1}} &0.516&0.584&0.312&96.744&94.847&93.348\\

         \midrule[0.5pt]
        \multirow{1}{*}{NSFormer, \shortstack{\textit{r}=0.2}} &0.518&0.586&0.312&97.481&95.427&93.664\\

       \midrule[0.5pt]
        \multirow{1}{*}{NSFormer, \shortstack{\textit{r}=0.3}}   &0.519&0.582&0.314&98.354&96.752&94.328\\
          \midrule[0.5pt]
        \multirow{1}{*}{NSFormer, \shortstack{\textit{r}=0.4}}  &0.517&0.582&0.314&98.932&97.242&95.423\\
      \midrule[0.5pt]
    \end{tabular}}
\vspace{-0.4cm}
\end{table}

\begin{table*}[bt]
    \setlength{\tabcolsep}{10pt}
    % {|>{\setlength{\tabcolsep}{3pt}}c|c|c|}
    \centering
    %% \vspace{-0.2cm}
    \caption{Full results of CATS, ModernTCN, and TimeXer on long-term TSF on popular public datasets with average performance on four random seeds.  Due to space limitations, the standard deviation is omitted, and we will show it in our full version paper. }
    %% \vspace{-0.2cm}
    \label{tab:full_longterm_p0}
    % \begin{tabular}{c|c|p{20pt}p{20pt}|cc|cc|cc|cc|cc}
    { \footnotesize
    \begin{tabular}{c|c|cc|cc|cc|cc|cc|cc}
        \hline
        \multirow{2}{*}{\shortstack{}} & &  \multicolumn{2}{c|}{CATS-APTF} &  \multicolumn{2}{c|}{CATS}&  \multicolumn{2}{c|}{ModernTCN-APTF} &  \multicolumn{2}{c|}{ModernTCN} &  \multicolumn{2}{c|}{TimeXer-APTF} &  \multicolumn{2}{c}{TimeXer} \\
         & & MSE & MAE & MSE & MAE & MSE & MAE & MSE & MAE & MSE & MAE & MSE & MAE    \\ 
        \midrule[0.5pt]
         \multirow{4}{*}{\rotatebox[origin=c]{90}{ETTh1}} 
&96 &\textbf{0.379} &\textbf{0.4}&0.389 &0.408&\textbf{0.37} &\textbf{0.399}&0.381 &0.411&\textbf{0.394} &\textbf{0.412}&0.396 &0.416\\
&192 &\textbf{0.4} &\textbf{0.417}&0.409 &0.424&\textbf{0.403} &\textbf{0.422}&0.42 &0.439&\textbf{0.421} &\textbf{0.431}&0.423 &0.438\\
&336 &\textbf{0.421} &\textbf{0.434}&0.422 &0.436&\textbf{0.428} &\textbf{0.438}&0.441 &0.452&\textbf{0.436} &\textbf{0.448}&0.441 &0.459\\
&720 &\textbf{0.482} &0.473&\textbf{0.482} &\textbf{0.471}&\textbf{0.47} &\textbf{0.477}&0.476 &0.481&\textbf{0.462} &\textbf{0.485}&0.488 &0.498\\
        \midrule[0.5pt]
              \multirow{4}{*}{\rotatebox[origin=c]{90}{ETTh2}} 
&96 &\textbf{0.278} &\textbf{0.345}&0.28 &0.35&\textbf{0.269} &\textbf{0.335}&0.279 &0.347&\textbf{0.278} &\textbf{0.346}&0.279 &0.347\\
&192 &\textbf{0.331} &\textbf{0.38}&0.334 &0.389&\textbf{0.33} &\textbf{0.374}&0.333 &0.384&\textbf{0.34} &\textbf{0.383}&0.348 &0.39\\
&336 &\textbf{0.354} &\textbf{0.402}&0.36 &0.41&\textbf{0.365} &\textbf{0.407}&0.369 &0.412&\textbf{0.363} &\textbf{0.405}&0.366 &0.41\\
&720 &\textbf{0.386} &\textbf{0.428}&0.413 &0.448&\textbf{0.403} &\textbf{0.435}&0.419 &0.449&\textbf{0.392} &\textbf{0.429}&0.406 &0.443\\
\midrule[0.5pt]
                 \multirow{4}{*}{\rotatebox[origin=c]{90}{ETTm1}} 
&96 &\textbf{0.29} &\textbf{0.34}&0.299 &0.352&\textbf{0.298} &\textbf{0.347}&0.308 &0.36&\textbf{0.309} &\textbf{0.357}&0.315 &0.363\\
&192 &\textbf{0.319} &\textbf{0.359}&0.327 &0.368&\textbf{0.346} &\textbf{0.375}&0.354 &0.382&\textbf{0.35} &\textbf{0.38}&0.353 &0.385\\
&336 &\textbf{0.351} &\textbf{0.378}&0.358 &0.387&\textbf{0.374} &\textbf{0.388}&0.393 &0.405&\textbf{0.382} &\textbf{0.4}&\textbf{0.382} &0.404\\
&720 &\textbf{0.406} &\textbf{0.413}&0.418 &0.423&\textbf{0.418} &\textbf{0.421}&0.446 &0.44&\textbf{0.414} &\textbf{0.425}&0.425 &0.433\\
\midrule[0.5pt]
                         \multirow{4}{*}{\rotatebox[origin=c]{90}{ETTm2}} 
&96 &\textbf{0.165} &\textbf{0.254}&0.189 &0.283&\textbf{0.161} &\textbf{0.249}&0.169 &0.26&\textbf{0.165} &\textbf{0.254}&0.174 &0.262\\
&192 &\textbf{0.219} &\textbf{0.293}&0.245 &0.321&\textbf{0.217} &\textbf{0.288}&0.226 &0.301&\textbf{0.222} &\textbf{0.294}&0.236 &0.306\\
&336 &\textbf{0.274} &\textbf{0.33}&0.302 &0.356&\textbf{0.28} &\textbf{0.329}&0.292 &0.342&\textbf{0.272} &\textbf{0.326}&0.287 &0.338\\
&720 &\textbf{0.374} &\textbf{0.398}&0.393 &0.413&\textbf{0.392} &\textbf{0.397}&0.414 &0.412&\textbf{0.357} &\textbf{0.38}&0.367 &0.391\\
\midrule[0.5pt]
                         \multirow{4}{*}{\rotatebox[origin=c]{90}{Weather}} 
&96 &\textbf{0.147} &\textbf{0.202}&0.148 &0.205&\textbf{0.142} &\textbf{0.191}&0.151 &0.205&\textbf{0.152} &\textbf{0.204}&0.156 &0.209\\
&192 &\textbf{0.191} &\textbf{0.244}&0.192 &0.246&\textbf{0.192} &\textbf{0.241}&0.196 &0.248&\textbf{0.199} &\textbf{0.246}&\textbf{0.199} &0.247\\
&336 &\textbf{0.242} &\textbf{0.283}&0.243 &0.285&\textbf{0.247} &\textbf{0.285}&0.249 &0.288&\textbf{0.248} &\textbf{0.284}&0.249 &0.285\\
&720 &\textbf{0.316} &\textbf{0.336}&\textbf{0.316} &\textbf{0.336}&\textbf{0.328} &\textbf{0.34}&0.331 &0.342&0.314 &\textbf{0.333}&\textbf{0.313} &0.334\\
\midrule[0.5pt]
                                 \multirow{4}{*}{\rotatebox[origin=c]{90}{Exchange}} 
&96 &\textbf{0.083} &\textbf{0.202}&0.084 &0.203&\textbf{0.082} &\textbf{0.199}&0.085 &0.203&\textbf{0.087} &\textbf{0.207}&\textbf{0.087} &0.208\\
&192 &\textbf{0.17} &\textbf{0.295}&0.175 &0.298&\textbf{0.168} &\textbf{0.292}&0.173 &0.296&\textbf{0.179} &\textbf{0.302}&0.181 &\textbf{0.302}\\
&336 &\textbf{0.314} &\textbf{0.404}&0.326 &0.412&\textbf{0.308} &\textbf{0.402}&0.325 &0.412&\textbf{0.326} &\textbf{0.414}&0.338 &0.42\\
&720 &\textbf{0.78} &\textbf{0.663}&0.825 &0.684&\textbf{0.823} &\textbf{0.682}&0.854 &0.698&\textbf{0.825} &\textbf{0.685}&0.865 &0.702\\
\midrule[0.5pt]
                   \multirow{4}{*}{\rotatebox[origin=c]{90}{Electricity}} 
&96 &\textbf{0.202} &\textbf{0.28}&0.209 &0.287&\textbf{0.152} &\textbf{0.255}&0.157 &0.263&\textbf{0.216} &\textbf{0.315}&0.221 &0.32\\
&192 &\textbf{0.215} &\textbf{0.292}&0.22 &0.297&\textbf{0.171} &\textbf{0.272}&0.181 &0.284&\textbf{0.219} &\textbf{0.318}&0.224 &0.323\\
&336 &\textbf{0.233} &\textbf{0.308}&0.236 &0.312&\textbf{0.189} &\textbf{0.291}&0.198 &0.3&\textbf{0.234} &\textbf{0.332}&0.24 &0.337\\
&720 &\textbf{0.281} &\textbf{0.349}&0.286 &0.355&\textbf{0.226} &\textbf{0.323}&0.234 &0.33&\textbf{0.277} &\textbf{0.363}&0.282 &0.368\\
\midrule[0.5pt]
                           \multirow{4}{*}{\rotatebox[origin=c]{90}{Traffic}} 
&96 &\textbf{0.758} &\textbf{0.456}&0.767 &0.465&\textbf{0.562} &\textbf{0.348}&0.579 &0.357&\textbf{0.698} &\textbf{0.439}&0.703 &0.442\\
&192 &\textbf{0.725} &\textbf{0.445}&0.741 &0.457&\textbf{0.596} &\textbf{0.374}&0.621 &0.388&\textbf{0.662} &\textbf{0.426}&0.669 &0.43\\
&336 &\textbf{0.734} &\textbf{0.448}&0.751 &0.46&\textbf{0.633} &\textbf{0.396}&0.665 &0.413&\textbf{0.665} &\textbf{0.422}&0.678 &0.432\\
&720 &\textbf{0.824} &\textbf{0.496}&0.855 &0.519&\textbf{0.726} &\textbf{0.437}&0.758 &0.45&\textbf{0.707} &\textbf{0.441}&0.714 &0.445\\
\hline

    \end{tabular}}
    % }
%% \vspace{-0.4cm}
\end{table*}

 \begin{table*}[h]
    \setlength{\tabcolsep}{3.5pt}
    % {|>{\setlength{\tabcolsep}{3pt}}c|c|c|}
    \centering
    %% \vspace{-0.2cm}
    \caption{Full results of long-term forecasting on remaining baselines.  with average performance on four random seeds. Due to space limitations, the standard deviation is omitted, and we will show it in our full version paper.}
    %% \vspace{-0.3cm}
    \label{tab:main_res_longterm_appendix}
    % \begin{tabular}{c|c|p{20pt}p{20pt}|cc|cc|cc|cc|cc}
    { \footnotesize
    \begin{tabular}{c|c|cc|cc|cc|cc|cc|cc|cc|cc}
        \hline
        \multirow{2}{*}{\shortstack{}} & &  \multicolumn{2}{c|}{TimeMixer} &  \multicolumn{2}{c|}{PatchTST}&  \multicolumn{2}{c|}{Autoformer} &  \multicolumn{2}{c|}{NSformer} &  \multicolumn{2}{c|}{Scaleformer} &  \multicolumn{2}{c|}{NLinear} &  \multicolumn{2}{c|}{N-HiTS} &  \multicolumn{2}{c}{Informer} \\
         & & +APTF & Original & +APTF & Original & +APTF & Original & +APTF & Original & +APTF & Original & +APTF & Original& +APTF & Original  & +APTF & Original \\ 
        \midrule[0.5pt]
         \multirow{4}{*}{\rotatebox[origin=c]{90}{ETTh1}} 
        &96 &\textbf{0.434} &0.45 &\textbf{0.377} &0.38 &\textbf{0.465} &0.492 &\textbf{0.512} &0.544 &\textbf{0.506} &0.507 &\textbf{0.423} &0.442 &\textbf{0.432} &0.448 &\textbf{0.9} &0.948 \\

         &192 &\textbf{0.49} &0.522 &\textbf{0.406} &0.412 &\textbf{0.479} &0.494 &\textbf{0.5} &0.51 &\textbf{0.494} &0.495 &\textbf{0.449} &0.464 &\textbf{0.466} &0.484 &\textbf{1.031} &1.05\\

         & 336 &\textbf{0.512} &0.535 &\textbf{0.427} &0.436 &\textbf{0.476} &0.484 &\textbf{0.49} &0.496 &\textbf{0.566} &0.579 &\textbf{0.466} &0.478 &\textbf{0.496} &0.516 &\textbf{1.079} &1.056\\

         & 720 &\textbf{0.532} &0.56 &\textbf{0.472} &0.49 &\textbf{0.548} &0.544 &\textbf{0.541} &0.542 &\textbf{0.607} &0.619 &\textbf{0.475} &0.492 &\textbf{0.57} &0.616 &\textbf{1.141} &1.176\\
        \midrule[0.5pt]
         \multirow{4}{*}{\rotatebox[origin=c]{90}{ETTh2}} 
        &96 &\textbf{0.316} &0.326 &\textbf{0.304} &0.31 &\textbf{0.376} &0.409 &\textbf{0.39} &0.406 &\textbf{0.381} &0.408 &\textbf{0.318} &0.328 &\textbf{0.349} &0.385 &\textbf{1.225} &1.568 \\

         &192 &\textbf{0.396} &0.43 &\textbf{0.357} &0.369 &\textbf{0.382} &0.412 &\textbf{0.396} &0.418 &\textbf{0.39} &0.429 &\textbf{0.372} &0.378 &\textbf{0.416} &0.470 &\textbf{1.471} &2.579\\

         & 336  &\textbf{0.43} &0.446 &\textbf{0.388} &0.386 &\textbf{0.386} &0.422 &\textbf{0.409} &0.428 &\textbf{0.4} &0.436 &0.403 &\textbf{0.401} &\textbf{0.568} &0.631 &\textbf{1.589} &2.718\\

         & 720 &\textbf{0.44} &0.457 &\textbf{0.414} &0.415 &\textbf{0.446} &0.494 &\textbf{0.434} &0.482 &\textbf{0.464} &0.492 &\textbf{0.426} &0.426 &\textbf{0.853} &0.91 &\textbf{2.024} &2.562\\
       \midrule[0.5pt]
         \multirow{4}{*}{\rotatebox[origin=c]{90}{ETTm1}} 
        &96 &\textbf{0.343} &0.356 &\textbf{0.308} &0.316 &\textbf{0.489} &0.502 &\textbf{0.519} &0.526 &0.484 &\textbf{0.478} &\textbf{0.334} &0.334 &\textbf{0.334} &0.348 &\textbf{0.514} &0.652 \\

         &192 &\textbf{0.385} &0.396 &\textbf{0.342} &0.355 &\textbf{0.526} &0.558 &\textbf{0.535} &0.566 &\textbf{0.514} &0.542 &\textbf{0.36} &0.362 &\textbf{0.371} &0.382 &\textbf{0.613} &0.707\\

         & 336 &\textbf{0.407} &0.414 &\textbf{0.369} &0.387 &\textbf{0.525} &0.605 &\textbf{0.571} &0.649 &\textbf{0.521} &0.528 &\textbf{0.382} &0.386 &\textbf{0.394} &0.41 &\textbf{0.712} &0.75 \\

         & 720 &\textbf{0.456} &0.461 &\textbf{0.41} &0.419 &\textbf{0.485} &0.559 &\textbf{0.588} &0.656 &\textbf{0.504} &0.526 &\textbf{0.423} &0.428 &\textbf{0.44} &0.456 &\textbf{0.861} &0.852\\
        \midrule[0.5pt]
         \multirow{4}{*}{\rotatebox[origin=c]{90}{ETTm2}} 
        &96 &\textbf{0.216} &0.221 &\textbf{0.202} &0.212 &\textbf{0.31} &0.318 &\textbf{0.282} &0.302 &\textbf{0.278} &0.293 &\textbf{0.21} &0.217 &\textbf{0.224} &0.244 &\textbf{0.888} &1.163 \\

         &192 &\textbf{0.27} &0.274 &\textbf{0.249} &0.264 &\textbf{0.332} &0.338 &\textbf{0.316} &0.33 &\textbf{0.343} &0.391 &\textbf{0.256} &0.262 &\textbf{0.277} &0.298 &\textbf{1.048} &1.204\\

         & 336 &\textbf{0.318} &0.323 &\textbf{0.294} &0.31 &\textbf{0.353} &0.362 &\textbf{0.344} &0.361 &\textbf{0.367} &0.463 &\textbf{0.302} &0.306 &\textbf{0.334} &0.356 &\textbf{1.692} &2.142\\

         & 720 &\textbf{0.395} &0.398 &\textbf{0.366} &0.375 &\textbf{0.408} &0.424 &\textbf{0.408} &0.432 &\textbf{0.428} &0.508 &\textbf{0.374} &0.378 &\textbf{0.408} &0.44 &\textbf{2.572} &3.857\\

        \midrule[0.5pt]
         \multirow{4}{*}{\rotatebox[origin=c]{90}{weather}} 
        &96 &\textbf{0.192} &0.196 &\textbf{0.169} &0.173 &\textbf{0.351} &0.357 &\textbf{0.246} &0.272 &\textbf{0.326} &0.372 &\textbf{0.198} &0.207 &\textbf{0.182} &0.188 &\textbf{0.316} &0.337 \\

         &192 &\textbf{0.235} &0.24 &\textbf{0.213} &0.216 &\textbf{0.37} &0.37 &\textbf{0.282} &0.288 &\textbf{0.359} &0.58 &\textbf{0.238} &0.246 &\textbf{0.229} &0.236 &\textbf{0.43} &0.436\\

         & 336 &\textbf{0.284} &0.29 &\textbf{0.262} &0.263 &\textbf{0.39} &0.397 &\textbf{0.319} &0.32 &\textbf{0.498} &0.559 &\textbf{0.279} &0.286 &\textbf{0.278} &0.282 &\textbf{0.53} &0.58\\

         & 720 &\textbf{0.346} &0.35 &0.325 &\textbf{0.324} &0.42 &\textbf{0.418} &0.382 &\textbf{0.374} &0.627 &\textbf{0.623} &\textbf{0.338} &0.34 &\textbf{0.34} &0.344 &\textbf{0.756} &1.027\\
         
       \midrule[0.5pt]
         \multirow{4}{*}{\rotatebox[origin=c]{90}{Exchange}} 
        &96&\textbf{0.146} &0.15&\textbf{0.142} &0.143&\textbf{0.206} &0.218&\textbf{0.2} &0.211&\textbf{0.2} &0.213&\textbf{0.144} &0.147&\textbf{0.168} &0.172&\textbf{0.582} &0.706 \\

         &192 &\textbf{0.237} &0.242&\textbf{0.23} &0.233&\textbf{0.306} &0.32&\textbf{0.308} &0.308&\textbf{0.302} &0.324&\textbf{0.232} &0.236&\textbf{0.251} &0.309&\textbf{0.816} &0.88\\

         & 336 &\textbf{0.359} &0.368&\textbf{0.355} &0.362&\textbf{0.446} &0.459&\textbf{0.436} &0.442&\textbf{0.455} &0.518&\textbf{0.357} &0.364&\textbf{0.386} &0.504&\textbf{1.002} &1.006\\

         & 720 &\textbf{0.742} &0.764&\textbf{0.741} &0.756&\textbf{0.863} &0.956&\textbf{0.831} &0.881&\textbf{1.01} &1.192&\textbf{0.746} &0.756&\textbf{0.619} &0.622&\textbf{1.099} &1.102\\

      \midrule[0.5pt]
         \multirow{4}{*}{\rotatebox[origin=c]{90}{Electricity}} 
        &96 &\textbf{0.262} &0.279&\textbf{0.23} &0.232&\textbf{0.304} &0.324&\textbf{0.282} &0.29&\textbf{0.288} &0.308&\textbf{0.302} &0.338&\textbf{0.232} &0.238&\textbf{0.588} &0.596 \\

         &192 &\textbf{0.272} &0.291&\textbf{0.234} &0.237&\textbf{0.327} &0.328&\textbf{0.308} &0.32&\textbf{0.318} &0.324&\textbf{0.304} &0.341&\textbf{0.239} &0.246&\textbf{0.708} &0.75\\

         & 336 &\textbf{0.29} &0.308&\textbf{0.249} &0.252&\textbf{0.334} &0.34&\textbf{0.331} &0.334&\textbf{0.348} &0.37&\textbf{0.318} &0.355&\textbf{0.256} &0.264&\textbf{0.804} &0.821\\

         & 720 &\textbf{0.328} &0.344&\textbf{0.286} &0.289&\textbf{0.376} &0.382&\textbf{0.363} &0.378&\textbf{0.39} &0.396&\textbf{0.352} &0.387&\textbf{0.292} &0.299&\textbf{0.85} &0.854\\

    \midrule[0.5pt]
         \multirow{4}{*}{\rotatebox[origin=c]{90}{Traffic}} 
        &96 &\textbf{0.619} &0.648&\textbf{0.512} &0.514&\textbf{0.59} &0.615&\textbf{0.568} &0.588&\textbf{0.563} &0.609&\textbf{0.714} &0.799&\textbf{0.472} &0.482&\textbf{0.736} &0.74 \\

         &192 &\textbf{0.594} &0.623&\textbf{0.479} &0.485&\textbf{0.64} &0.642&\textbf{0.659} &0.694&0.671 &\textbf{0.656}&\textbf{0.681} &0.778&\textbf{0.472} &0.486&\textbf{0.902} &0.913\\

         & 336 &\textbf{0.594} &0.633&\textbf{0.486} &0.49&\textbf{0.604} &0.638&\textbf{0.64} &0.668&\textbf{0.657} &0.692&\textbf{0.689} &0.788&\textbf{0.486} &0.501&\textbf{1.039} &1.112\\

         & 720 &\textbf{0.644} &0.687&\textbf{0.515} &0.517&\textbf{0.637} &0.674&\textbf{0.663} &0.692&\textbf{0.665} &0.702&\textbf{0.716} &0.817&\textbf{0.526} &0.542&\textbf{1.137} &1.138\\

\hline

    \end{tabular}}
    % }
%% \vspace{-0.1cm}
\end{table*}

\end{document}